\documentclass{article} % For LaTeX2e
\usepackage[preprint]{colm2026_conference}

\usepackage{listings}
\lstdefinelanguage{json}{
    basicstyle=\ttfamily\small,
    stringstyle=\color{red},
    commentstyle=\color{gray},
    morestring=[b]",
    morecomment=[l]{//},
}
\usepackage[T1]{fontenc}
\usepackage[utf8]{inputenc} % (only needed if using pdflatex)
\usepackage{microtype}
\usepackage{hyperref}
\usepackage{url}
\usepackage{booktabs}
\usepackage{graphicx}
\usepackage{amsmath}
\newcommand{\extrabold}[1]{%
  \textbf{\smash{\raisebox{0pt}{#1}}}%
  \llap{\textbf{\smash{\raisebox{0pt}{#1}}}}%
}

\usepackage{lineno}

\definecolor{darkblue}{rgb}{0, 0, 0.5}
\hypersetup{colorlinks=true, citecolor=darkblue, linkcolor=darkblue, urlcolor=darkblue}

\title{MTA-Agent: An Open Recipe for Multimodal Deep Search Agents}

\author{
  \hspace{-0.4cm}Xiangyu Peng\thanks{Equal contribution.} \hspace{0.15cm}
  Can Qin\footnotemark[1] \hspace{0.15cm}
  An Yan\thanks{Equal contribution.} \hspace{0.15cm}
  Xinyi Yang\footnotemark[2] \hspace{0.15cm}
  Zeyuan Chen \hspace{0.07cm}
  Ran Xu \hspace{0.07cm}
  Chien-Sheng Wu \\
  \hspace{5cm} Salesforce AI Research \\
  \hspace{2cm} \texttt{\{becky.peng, cqin, wu.jason\}@salesforce.com}
}

\usepackage{colortbl}

\usepackage{hyperref}
\usepackage{url}
\usepackage{booktabs}
\usepackage{multirow}
\usepackage{xcolor}
\usepackage{tikz}
\usepackage{enumitem}
\usepackage{algorithm}
\usepackage{algpseudocode}
\definecolor{brightcerulean}{rgb}{0.11, 0.67, 0.84}

\usepackage{threeparttable}

\fontsize{7pt}{8pt}\selectfont

\newcommand{\datasetname}{\texttt{MTA-Vision-DeepSearch}}
\newcommand{\framename}{\texttt{MTA-Agent}}
\newcommand{\modelnamebig}{\texttt{MTA-DeepSearch-32B}}
\newcommand{\modelnamesmall}{\texttt{MTA-DeepSearch-8B}}

\usepackage{longtable}
\usepackage{array} % for better column alignment
\usepackage{tcolorbox}

\begin{document}

\ifcolmsubmission
\linenumbers
\fi

\maketitle

\begin{abstract}
Multimodal large language models (MLLMs) have demonstrated strong capabilities in visual understanding, yet they remain limited in complex, multi-step reasoning that requires deep searching and integrating visual evidence with external knowledge. 
In this work, we address this challenge by constructing high-quality, verified multi-hop vision-language training data for multimodal deep-search agents. We propose a \textbf{M}ulti-hop \textbf{T}ool-\textbf{A}ugmented \textbf{Agent} for Evidence-based QA Synthesis (\framename), which automatically selects tools and their parameters to retrieve and validate evidence from both visual and textual sources and generates structured multi-hop question answer trajectories. 
Starting from diverse VQA seed datasets, our pipeline produces a large-scale training dataset, \datasetname, containing 21K high-quality multi-hop examples. 
The data is filtered through a multi-stage verification process to ensure factual consistency and answer uniqueness.
Using \datasetname, a 32B open-source multimodal search agent achieves state-of-the-art performance, reaching an average of $54.63\%$ across six challenging benchmarks, outperforming GPT-5 ($51.86\%$), Gemini-2.5-Pro ($50.98\%$), and Gemini-3-Pro ($54.46\%$) under the same tool settings. 
We further show that training on our data improves both reasoning depth and tool-use behavior, increasing the average number of steps from 2.27 to 4.28, and leading to more systematic and persistent search strategies. Additionally, we demonstrate that training can be performed without real-time tool calls by replaying cached interactions, significantly reducing training cost.
Importantly, we present \framename\ as a fully open recipe for multimodal deep search: we release the entire dataset, training trajectories, and implementation details to enable reproducibility and future research on open multimodal search agents.
Data and codes can be found in \href{https://github.com/SalesforceAIResearch/MTA-Agent}{https://github.com/SalesforceAIResearch/MTA-Agent}.
\end{abstract}

\section{Introduction}

Multimodal large language models (MLLMs) have achieved strong performance across a wide range of vision tasks~\citep{xue2024xgen,deepseekai2025deepseekv32,huang2025vision,qwen3technicalreport}. 
However, their performance is often limited by the parametric knowledge, making them struggle with complex, fact-intensive visual question answering (VQA) that requires up-to-date or long-tail real-world information~\citep{geng2025webwatcher,chng2025sensenova,wu2025mmsearch}. 
While proprietary systems have made progress by integrating ReAct-style~\citep{yao2022react} reasoning with tool use, multimodal deep search remains largely underexplored, with few existing agents capable of handling high-difficulty vision-language tasks~\citep{xu2025comprehensive,hu2025owl}.
We argue that the primary bottleneck is the lack of high-quality training data needed to transfer the strong VQA capabilities of MLLMs to more challenging deep-search scenarios.

Current approaches for transferring MLLMs into effective multimodal deep search
% \jw{research? search? need to be consistent in the paper} 
systems still face three major limitations.
First, existing open-source models used as the backbone of deep-search agents are constrained in both reasoning depth and search breadth~\citep{fu2025seeking,wu2025mmsearch,narayan2025deepmmsearch}. 
For example, before training, \texttt{Qwen3-VL-32B-Instruct}~\citep{qwen3technicalreport} performs only $1.76$ tool-use steps on average on the MMSearch-VL benchmark~\citep{jiang2024mmsearch}, achieving $68.52\%$ accuracy. In comparison, GPT-5 achieves $77.65\%$ accuracy with an average of $2.60$ search steps.
The absence of longer and more challenging training data for the underlying LLM backbone further limits the agent’s ability to solve questions that require aggregating evidence from multiple sources.
Second, the lack of high-quality training data remains a major bottleneck. Existing data synthesis approaches often produce inconsistencies between reasoning processes and final answers, which hinders effective learning~\citep{li2024benchmarking,wu2025mmsearch}. In contrast to VQA datasets, which have significantly advanced MLLM capabilities, high-quality datasets for multimodal deep search remain scarce.
Third, training data for deep search lacks diversity. Most datasets are constructed from single images with web search, resulting in limited coverage of complex and multi-step scenarios~\citep{geng2025webwatcher}. Such limited and homogeneous data encourages agents to rely on a single search strategy, restricting their ability to generalize to more complex reasoning tasks.

To address these limitations, we propose a multi-hop, tool-augmented agent for evidence QA synthesis framework (\textbf{\framename}). Unlike existing multimodal deep
% \jw{search?} 
search methods that often rely on shallow retrieval or limited reasoning patterns, our approach explicitly targets long-horizon, multi-step reasoning grounded in both visual and textual evidence.
Our key idea is to transform existing high-quality VQA resources into challenging multi-hop search tasks, enabling richer reasoning scenarios and more diverse tool usage. This allows us to preserve strong visual understanding while significantly increasing the complexity of the reasoning process.
We further introduce a tool-augmented agent that performs iterative search and reasoning, leveraging multiple tools and flexible interaction strategies rather than relying on a single retrieval pattern. This design encourages more robust and diverse exploration during evidence collection.
To ensure reliability, we incorporate a verification pipeline that filters out inconsistent or shortcut reasoning, resulting in high-quality multi-hop reasoning trajectories.
Finally, using the synthesized dataset (\textbf{\datasetname}), we train a single 32B dense model with RL, achieving state-of-the-art performance across six deep-search 
% \jw{search?} 
benchmarks and surpassing strong commercial search agents. Detailed technical design is provided in Section~\ref{sec:data}.

The main contributions of this work are as follows:
\begin{enumerate}[noitemsep,topsep=0pt,itemsep=0pt, leftmargin=*]
    \item We propose a multi-hop tool-augmented evidence QA synthesis agent (\framename) as a fully open recipe for multimodal deep search (\S\ref{sec:data}), which automatically generates high-quality multi-hop QA data from existing VQA datasets without human efforts.
    
    \item We release a $21$K high-quality training dataset (\datasetname) for multimodal deep-search agents and human-verified and rewritten test set for evaluation (\S~\ref{sec:data_stats}).
    
    \item We show that our dataset enables a 32B open-source model to achieve state-of-the-art performance, even outperforming leading commercial models (\S\ref{sec:exp}). We also release rollout histories, which enable tool-free training and further improve performance compared to training with real tool calls on other existing training dataset (\S\ref{sec:cache_train}).
    
    \item We conduct extensive analysis (\S\ref{sec:analysis}), including the impact of training data choices, strategies to improve data quality, and changes in tool usage behavior during training.
\end{enumerate}
\vspace{-2mm}
\section{Related Works}
\vspace{-2mm}

\textbf{Deep Search Agents.}
Recent advancements in LLM agents mark a shift from simple reactive or retrieval-based chatbots~\citep{chen2024mllm,yu2024visrag} to autonomous, goal-directed systems that can plan, self-correct, and execute complex workflows. To support this increased autonomy, modern agents are tightly integrated with tool-use capabilities and secure execution environments~\citep{openai2025deepresearch,wu2026deepresearch,li2025search}. 
\citet{song2025r1} propose R1-Searcher, a two-stage outcome-based RL approach to enhance search capabilities. 
\citet{wu2026deepresearch} develop DeepResearch-R1, an open-source training framework that supports multi-turn web interactions and diverse reward models, including rule-based outcomes and LLM-as-judge feedback. 
\citet{tao2025webshaper} introduce a formalization-driven data synthesis framework that enables precise control over reasoning structures through knowledge projectionoperations.
Despite these advances, most deep-search agents~\citep{yao2026mm} remain primarily text-centric. MMSearch-R1~\citep{wu2025mmsearch} extends search to multimodal settings by enabling dynamic image and text retrieval, while \citet{geng2025webwatcher} improve generalization through synthetic trajectories.
However, training data creation for these systems often requires substantial human effort, and many works do not release their original training datasets. 
As a result, existing training data remains limited in scale and diversity, restricting the development of multimodal deep search agents that can learn diverse and robust search strategies.

\textbf{Multimodal Deep Search Data.}
Most existing VQA datasets primarily evaluate single-step reasoning or shallow retrieval capabilities. For example, OK-VQA~\citep{okvqa} requires commonsense reasoning over everyday scenes, while MMT-Bench~\citep{ying2024mmt} focuses on expert-level knowledge and fine-grained visual understanding.
More recent datasets aim to improve and evaluate deep-search capabilities. LiveVQA~\citep{fu2025seeking} is designed to test reasoning over up-to-date, real-world visual information, and FVQA~\citep{wu2025mmsearch} requires the use of external knowledge. However, these tasks are relatively simple and can often be solved with only one or two search steps.
Recent efforts~\citep{li2024benchmarking,chen2025detecting,huang2026mmdeepresearch,zeng2026vision} such as MMSearch~\citep{jiang2024mmsearch} and MMSearch-Plus~\citep{tao2025mmsearch} provide more challenging benchmarks for multimodal search. 
However, they are primarily designed for evaluation and do not offer large-scale training data.
To address these limitations, we propose a data generation agent that interacts with four tools to construct single-hop QA pairs, which are then composed into multi-hop questions through a carefully designed verification process. Our approach produces high-quality training data that is scalable and allows explicit control over task difficulty and domain diversity.

\vspace{-3mm}
\section{Data Creation}
\vspace{-2mm}
\label{sec:data}

In this section, we describe how we construct a QA agent to generate training data. We first filter existing VQA datasets to serve as seeds (\S\ref{sec:seed}). 
Next, we design a QA agent (\S\ref{sec:agent}) that selects tools and their parameters to collect feedback and generate single-hop question answer pairs. 
We further implement a validation system to ensure that the synthetic QA data is factual, unique, and high-quality for deep search.
These single-hop questions are connected through entities and merged into multi-hop questions, which are used to train multimodal deep-search agents.

\vspace{-2mm}
\subsection{VQA Seed Selection}
\vspace{-1mm}
\label{sec:seed}

We use the training splits of FVQA~\citep{wu2025mmsearch}, LiveVQA News~\citep{fu2025seeking}, InfoVQA~\citep{mathew2022infographicvqa}, InfoSeek~\citep{chen2023can}, and OK-VQA~\citep{okvqa} as seed datasets. FVQA and LiveVQA contain news images and represent relatively easy deep-search settings. InfoVQA requires OCR over infographic images, InfoSeek involves Wikipedia images and external knowledge retrieval, and OK-VQA requires commonsense reasoning over complex everyday scenes. 
Ablation studies for the design choices are presented in Section~\ref{sec:data_chocie}.

Each VQA sample is processed through a multi-stage filtering pipeline. 
First, GPT-5-mini verifies that the question requires visual information, discarding those answerable from general knowledge. 
Second, GPT-5 rewrites the question into a clean, image-grounded free-form format and adds a short disambiguating entity description (e.g., ``Birmingham, city in Alabama''). Questions that cannot be converted from multiple-choice are rejected. 
Third, GPT-5 ensures the answer is a specific proper-noun entity from predefined categories, filtering out ambiguous or generic terms. 
An \textit{entity} is defined as a uniquely named proper noun (e.g., person, organization, location, product, event, or creative work), rather than a description, number, or multi-part answer (see Appendix~\ref{app:entity_def}).
Finally, GPT-5 with vision and web search verifies factual correctness; samples without a definitive answer are discarded. 
Full prompts and predefined categories are provided in Appendix~\ref{app:seed}.

% \vspace{-1mm}
\label{sec:agent}

\begin{figure}[htb!]
    % \centering
    \includegraphics[width=\linewidth]{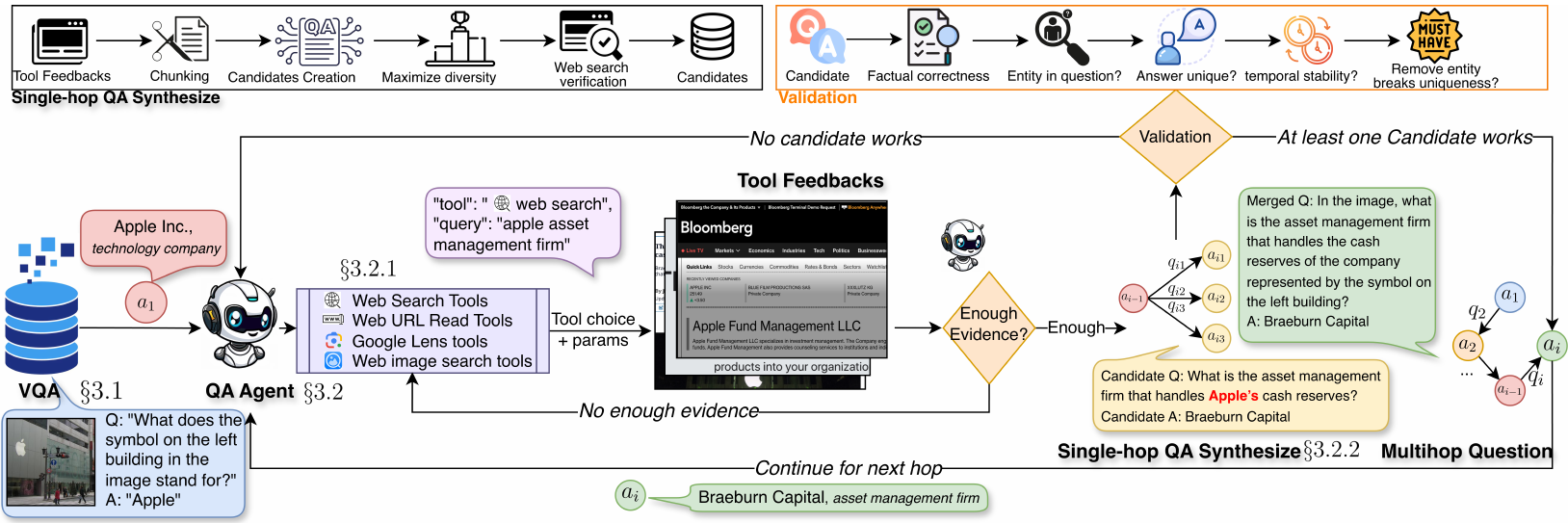}
    \vspace{-5mm}
\caption{Data creation pipeline. Starting from filtered seed VQA data, the QA agent iteratively selects tools to collect sufficient feedback for generating single-hop QA candidates. These candidates are verified, and the most diverse valid candidate is appended to the multi-hop reasoning chain. If no candidate passes verification, the agent continues to gather additional evidence by selecting tools. Detailed examples are shown in Appendix~\ref{app:examples}.}
    \label{fig:pipeline}
    \vspace{-3mm}
\end{figure}

\vspace{-2mm}
\subsection{QA Agent}
\vspace{-1mm}
% \begin{algorithm}[t]
% \caption{QA Agent for Multi-hop Question Construction}
% \label{alg:qa_agent}
% \begin{algorithmic}[1]

% \Require Seed VQA $(I, q_1, a_1)$, hop limit $K$
% \State Initialize chain $Q \gets [q_1]$, $A \gets [a_1]$

% \For{$k = 2$ to $K$}

%     \State \textbf{(1) Evidence Collection}
%     \State $\mathcal{E} \gets \emptyset$, $t \gets 0$
%     \While{$t < 4$ \textbf{or} insufficient evidence (agent decision)}
%         \State Execute one tool:
%         \State \quad \texttt{web\_search(query),} \texttt{web\_reader(url)}, 
%         \State \quad 
%         \texttt{image\_search(query)}, \texttt{google\_lens\_search(image\_url)}
%         \State Update $\mathcal{E}$, $t \gets t + 1$
%     \EndWhile

%     \State \textbf{(2) QA Candidate Generation}
%     \State Generate candidates $(q, a, c)$ using GPT-5.1 with $\mathcal{E}$
%     \State Select $a_k$ maximizing diversity from $A$ (Sentence-BERT)

%     \State \textbf{(3) Verification}
%     \If{$(q_k, a_k)$ fails verification}
%         \State \textbf{continue}
%     \EndIf

%     \State $\tilde{a}_k \gets \text{Disambiguate}(q_k, a_k, c_k)$

%     \State \textbf{(4) Chain Construction}
%     \State $\tilde{q}_k \gets \text{Merge}(Q, q_k)$

%     \State \textbf{(5) Anti-leakage Check}
%     \If{$\tilde{q}_k$ directly reveals the final answer}
%         \State \textbf{continue}
%     \EndIf

%     \State Append $\tilde{q}_k$ to $Q$, $\tilde{a}_k$ to $A$

% \EndFor

% \State \Return multi-hop questions $Q$ and answers $A$

% \end{algorithmic}
% \end{algorithm}

% As illustrated in Figure~\ref{fig:pipeline}, w
We design a QA agent with two types of actions: tool use (\S\ref{sec:tool_use}) and QA generation (\S\ref{sec:qa}). 
The agent follows a \textbf{ReAct} framework~\citep{yao2022react}, where it first generates reasoning and then selects an action from either tool use or QA generation. If it selects tool use, it further chooses a specific tool along with its parameters (e.g., search queries or image URLs). If it selects QA generation, it triggers the process described in \S~\ref{sec:qa}.
The agent conditions on the full interaction history, including prior reasoning steps, tool selections, tool feedback, and previously generated questions.

As shown in Figure~\ref{fig:pipeline}, the agent operates as follows. 
(1) It starts from a seed VQA instance consisting of an image, a question $q_1$, and an entity answer $a_1$.  
(2) The agent iteratively selects and uses tools to gather additional evidence (\S\ref{sec:tool_use}). 
(3) Once the QA agent determines that sufficient evidence has been collected, it will invokes the QA generation module (\S\ref{sec:qa}) by  to produce a new single-hop QA pair $(q_k, a_k)$. 
If the generated QA pair passes validation as described in Section~\ref{sec:qa}, the new question is merged with the existing question chain to form an extended multi-hop question $\tilde{q}_k$. 
If the QA pair fails validation, the failure reason is summarized and used to guide subsequent tool calls, with modified queries or URLs to collect additional evidence.
This process repeats iteratively, gradually extending the reasoning chain until a $K$-hop question is constructed.
Examples are shown in Appendix~\ref{app:examples}.

\vspace{-2mm}
\subsubsection{Tool Use for QA Generation}
\vspace{-1mm}
\label{sec:tool_use}

The agent is equipped with four tools:
(1) \textbf{web search}, which uses Tavily~\footnote{\href{https://www.tavily.com}{https://www.tavily.com}} to retrieve up to 10 URLs with titles and snippets given a text query; 
(2) \textbf{web reader}, which retrieves the full content of a given URL; 
(3) \textbf{Google Lens}, which searches for visually similar images and returns matched results with titles and snippets given an image; 
(4) \textbf{image search}, which retrieves image URLs given a text query.
We do not include OCR tools, as this capability is already covered by the VQA seed data. Additional details are provided in Appendix~\ref{app:tools}.
% \jw{do we want to add a motivation why these four, but not more others?}

\vspace{-1mm}
\subsubsection{Single-hop QA Creation Module}
\vspace{-1mm}
\label{sec:qa}

For each hop $k \in \{2, \dots, K\}$, given the bridge entity $a_{k-1}$ (the answer from the previous hop), we generate a new question $q_k$ and answer $a_k$ through a four-stage procedure. Ablation studies of this procedure are presented in Appendix~\ref{app:ablation_data}.

\textbf{Candidate Generation.}
The web content associated with $a_{k-1}$ is segmented into chunks of at most $5000$ tokens. Each chunk is scored based on language quality (e.g., sentence completeness; see Appendix~\ref{app:chunking}) and whether it contains $a_{k-1}$. We retain the top sources containing relevant information.
For each source, GPT-5.1 extracts a candidate triple $(q_k, a_k, c_k)$ (see Appendix~\ref{app:qa_gen}), where $c_k$ is a supporting excerpt. The prompt enforces that: 
(1) $q_k$ explicitly includes $a_{k-1}$; 
(2) $a_k$ is a unique entity distinct from $\{a_1, \ldots, a_{k-1}\}$; 
(3) $q_k$ is standalone; 
(4) removing $a_{k-1}$ makes the answer indeterminate; 
(5) $q_k$ has a single correct answer; and 
(6) $q_k$ captures a meaningful relation or attribute.
Each candidate is then verified for factual correctness against $c_k$, answer uniqueness, temporal stability, and entity dependency; invalid candidates are discarded. Further details are in Appendix~\ref{app:qa_synthesis}.

\textbf{Verification.}
The selected pair $(q_k,a_k)$ is verified by GPT-5 with live web search against five criteria: (1) factual correctness with respect to the source content, (2) explicit presence of $e_{k-1}$ in $q_k$, (3) unique answerability, (4) temporal stability of $a_k$, and (5) entity dependency (removing $e_{k-1}$ breaks uniqueness). The candidate is accepted only if all criteria are satisfied; otherwise it is discarded and generation continues with other chunks. Using a search-augmented verifier (GPT-5 with web search tools) provides an independent factual check beyond the generation source.
More details can be found in Appendix~\ref{app:qa_verification}.

\textbf{Answer Diversity Selection.}
When multiple candidates exist, we select the one that produces the most
challenging merged question. For each candidate, Qwen3-VL-32B~\citep{qwen3technicalreport} is asked what
search query ($r_\text{weak}$) it would use to answer the merged question $\tilde{q}_k$. The
weak model's query $r_\text{weak}$ is compared against the actual retrieval
query $r_\text{actual}$ via token-overlap Jaccard similarity (after stop-word
removal):
\[
  \text{sim}(r_\text{weak},\, r_\text{actual})
  = \frac{|\mathcal{T}(r_\text{weak}) \cap \mathcal{T}(r_\text{actual})|}
         {|\mathcal{T}(r_\text{weak}) \cup \mathcal{T}(r_\text{actual})|}
\]
We select the candidate with the lowest similarity score; if all candidates
exceed a threshold of $0.6$, the hop is rejected and a new search is
attempted. Answer diversity across hops is enforced separately: the synthesis
prompt lists all previous answers $\{a_1,\ldots,a_{k-1}\}$ and instructs the
model to produce a distinct answer. More details are in Appendix~\ref{app:candidate_selection}.

% When multiple candidates exist, we select the answer most semantically distinct from previously collected answers $\{a_1,\dots,a_{k-1}\}$. Using Sentence-BERT embeddings~\citep{reimers-2019-sentence-bert}, we apply the minimax criterion:
% \[
% a_k^*=\arg\min_{a\in C}\max_{j<k}\cos\big(\phi(a),\phi(a_j)\big),
% \]
% where $\phi(\cdot)$ denotes the encoder and $C$ is the candidate set. This prevents the answer chain from collapsing into semantically redundant entities. 
% For example, if a previous answer ($a_{i-1}$) is \textit{``Apple''}, a candidate  such as $a_{i1}$, \textit{``Braeburn Capital''} is preferred over $a_{i2}$, \textit{``Apple Music''}, as it is less semantically similar and introduces more diverse information.

\textbf{Answer Disambiguation.}
Because short entity answers (e.g., \textit{Apple}) may be ambiguous for the content search for next hop, we append a short disambiguating phrase using GPT-5.1.
The model adds a short phrase from the supporting context $c_k$ that uniquely identify the entity (e.g., \textit{Apple Inc., a technology company}). The disambiguated form $\tilde{a}_k$ is used as the seed for search query for hop $k{+}1$, while the original $a_k$ remains the ground-truth answer label. See more details in Appendix~\ref{app:disambiguation}.

\textbf{Anti-leakage Check.}
After merging $q_k$ with the existing reasoning chain to form the intermediate multi-hop question, we perform an anti-leakage test using Tavily. 
For example, consider the question: \textit{``Where is Donald Trump's hometown in the capital of the country shown in the image?''} The country (the United States) can be easily inferred.
Such cases are filtered out to prevent information leakage and to ensure that the final questions require genuine multi-hop reasoning and image reading.
Specifically, we query Tavily with the merged question and check whether the system can directly retrieve the final answer. If the answer can be obtained, the sample is discarded and generation continues with alternative candidates.

\vspace{-2mm}
\subsection{Data Statistics}
\label{sec:data_stats}
\vspace{-1mm}

\begin{table}[tb!]
\small
\setlength\tabcolsep{8pt}
\centering
\vspace{-1mm}
\begin{tabular}{l|rrrrr|r||rr}
\toprule
\textbf{Seed Dataset} & \textbf{Original} & \textbf{2-hop} & \textbf{3-hop} & \textbf{4-hop} & \textbf{Total} & \textbf{Test} & \textbf{Cost} & \textbf{Tools}\\
\midrule
FVQA     & 4,750 & 1,155 & 1,153 & 1,113 & 8,171 & 33 & \$ 0.30 & 9.8\\
Infoseek & 1,942 & 1,135 & 1,130 &   934 & 5,141 & 51 & \$ 0.25 & 8.4\\
InfoVQA  &   217 &   483 &   481 &   400 & 1,581 & 38 & \$ 0.28 & 9.8\\
News     &   918 &   240 &   238 &   236 & 1,632 & 13 & \$ 0.27 & 8.6\\
OK-VQA   &   629 & 1,418 & 1,415 & 1,168 & 4,630  & 43& \$ 0.31 & 10.1\\
\midrule
Total    & 8,456 & 4,431 & 4,417 & 3,851 & 21,155 & 178 & \$ 0.28 & 9.3\\
\bottomrule
\end{tabular}
\caption{Number of scenarios per hop in our training and test data. "Cost" and ``Tools" indicate the average LLM cost and number of tool calls per sample for generation.}
\label{tab:stats}
\vspace{-3mm}
\end{table}

For each reasoning chain, we retain all intermediate question--answer pairs, including 2-hop, 3-hop, and 4-hop cases. This allows the model to progressively learn from simpler to more complex reasoning tasks. Ablation studies for this design choice are presented in Section~\ref{sec:data_chocie}.
Table~\ref{tab:stats} summarizes the training data statistics. In total, the dataset contains 21K training examples. The ground-truth answers are short and can be reliably verified by a small language model.
We also construct a test set of $178$ examples, which are manually verified and rewritten by human annotators to evaluate the capability of deep-search agents. The average LLM cost per sample is $\$0.28$, with an average of $9.3$ tool calls per multi-hop QA generation.
Detailed examples are shown in Appendix~\ref{app:examples}.

\textbf{Human verification}
We conduct human evaluation to verify the quality of our generated training data, based on $444$ multi-hop trajectories. Each trajectory is annotated by three human annotators at both the single-hop and merged multi-hop levels.
We evaluate the data along three dimensions:
(1) \textbf{Understandability}: whether the final multi-hop question is clear and understandable to humans. Annotators find that $97.7\%$ of the questions are clearly understandable.
(2) \textbf{Factuality}: whether each single-hop QA pair is factually correct. $89.4\%$ of the hop-level answers are verified as correct.
(3) \textbf{Image requirement}: whether the question requires visual information to answer. $76.3\%$ of the questions are verified to require the image, meaning the answer cannot be determined without visual input. 
Although a subset of questions does not strictly require visual input, these examples are still useful for training, as they help improve the model’s general search and reasoning capabilities.
See more details about human evaluation in Appendix~\ref{app:human_validation}

\vspace{-3mm}
\section{Experiments}
\label{sec:exp}
\vspace{-1mm}
% In this section, we present a comprehensive experimental study.

% First, we compare our approach with existing methods on a range of multimodal web search reasoning benchmarks. 
% We also conduct ablation studies to examine the contribution of different components. 
% Our results show that supervised fine-tuning (SFT) does not improve performance and, in some cases, even leads to performance degradation. In addition, incorporating a reflection module does not provide further performance gains.
% Finally, we report the model's performance when using cached results instead of calling real APIs, demonstrating its effectiveness under this setting as well.
% Finally, we analyze the number of tool calls before and after training to understand how training affects tool usage behavior.

\vspace{-1mm}
\subsection{Experiment Setup}
\vspace{-1mm}
\subsubsection{Evaluation Datasets}
\vspace{-1mm}
We evaluate our approach and all baselines on six deep-search-oriented benchmarks, which represent the current state-of-the-art in multimodal search evaluation: MMSearch~\citep{jiang2024mmsearch}, MMSearch-Plus~\citep{tao2025mmsearch}, BrowseComp-VL (BC-VL)~\citep{geng2025webwatcher}, HR-MMSearch~\citep{chng2025sensenova}, the FVQA test split~\citep{wu2025mmsearch}, and our newly constructed test dataset, \datasetname-test (MTA-test).

\vspace{-2mm}
\subsubsection{Search Agent Setup}
\label{sec:agent_setup}
\vspace{-1mm}

We adopt a \textbf{ReAct} (Reasoning and Acting) framework~\citep{yao2022react} for our multimodal deep search agent.
At each step, the agent receives the research question alongside the input image, reasons over its current knowledge state, selects a tool to execute, and incorporates the resulting observation before deciding on the next action (Check Appendix~\ref{app:search_agent}). 

\textbf{ReAct Loop.}
The agent operates for up to $T=6$ iterations.
At each iteration, the agent produces a structured response containing:
(i) a \textit{reasoning} trace over the current state,
(ii) an \textit{action} specifying a tool and its parameters,
(iii) a binary \texttt{should\_stop} flag, and
(iv) a scalar \textit{confidence} score.
The loop terminates early if \texttt{should\_stop=true} with confidence above $0.7$, or if two consecutive tool calls fail.
Upon termination, the agent generates a final answer conditioned on the full dialogue history. More details can be found in Appendix~\ref{app:react}.

\textbf{Tools.}
The agent has access to seven tools:
\textit{Web search} retrieves ranked web snippets, titles, and URLs.
\textit{Web image search} returns text descriptions of image results from the web.
\textit{Web URL reader} extracts the full text content of a given webpage.
\textit{Reverse image search} leverages Google Lens to identify objects and landmarks and retrieve visually similar images.
\textit{OCR} extracts all visible text from the image.
\textit{Python execution} runs code in a persistent IPython session for data analysis and computation.
% \textit{Bash terminal} executes shell commands for file and system operations.
All tool calls are dispatched to a local observation server and the returned text observation is appended to the dialogue history as a user turn and fed back to the agent in the next iteration. Details are in Appendix~\ref{app:tool_agent}.
Note that tools not used in the training QA generation pipeline (Section~\ref{sec:data}) are included to support downstream question answering and out-of-distribution questions, but are not required during QA construction.
% \jw{I got confused how do you resolve different tools for training QA generation (4) and testing (7). Can you explain this here? }

\vspace{-1mm}
\subsubsection{Baselines}
\vspace{-1mm}
We compare our fine-tuned model with three commercial large language models: GPT-5~\citep{singh2025openai}, Gemini-2.5-Pro and Gemini-3-Pro~\citep{gemini3_2025}. For each model, we evaluate two settings: (1) direct answering without tools, and (2) the same agent framework with the identical tool setup used in our approach.
We also compare against two open-source models, Qwen3-VL-8B and Qwen3-VL-32B, evaluated both with and without the same agent tool configuration.
In addition, we report benchmark results from prior work. 
However, different papers adopt different tool setups, we directly list the numbers they reported in their original papers.
% \jw{should we mark a * next to the model name? like MMSearch-R1-7B* in Table 2, so people know which row we ran it which row we copied}. 
Specifically, we compare with SenseNova-MARS~\citep{chng2025sensenova}, WebWatcher~\citep{geng2025webwatcher}, MM-DeepResearch 32B~\citep{yao2026mm} and MMSearch-R1-7B~\citep{wu2025mmsearch}.

\vspace{-2mm}
\subsubsection{Training Setup}
\vspace{-1mm}
We fine-tune two models, Qwen3-VL-32B-Instruct and Qwen3-VL-8B-Instruct~\citep{qwen3technicalreport}, using the training data summarized in Table~\ref{tab:stats}, resulting in two models: \modelnamebig~and \modelnamesmall. 
Training is performed with DAPO~\citep{yu2025dapo}, using a batch size of 64, 8 rollout samples per prompt, a learning rate of $2\times10^{-6}$, and 100 training steps.
Each rollout uses a ReAct-style agent that interacts with tools for up to 6 steps. Observation tokens are masked from the policy gradient loss. Rewards are computed asynchronously using an LLM-based judge (see Appendix~\ref{app:llm-judge}.).
The 32B and 8B models are trained on 32 and 8 H100 GPUs, respectively, using the \texttt{verl-tool} framework\footnote{\url{https://github.com/TIGER-AI-Lab/verl-tool}}.

\vspace{-1mm}
\subsection{Main Results}
% \vspace{-1mm}

\begin{table*}[t]
    \centering
    \small
    \setlength\tabcolsep{1.5pt}
    \begin{threeparttable}
    \begin{tabular}{l|cccccc|c}
    \toprule
    Model
    & MMSrch-Plus
    & HR-MMSrch
    & BC-VL 
    & MMSrch
    & FVQA
    & MTA-test
    & Avg\\
    \midrule
    \multicolumn{8}{c}{\hspace{0.5cm}\textbf{Direct Answer}}  \\
    \midrule
    GPT-5 & 13.18  & 16.39 & 41.60 & 36.47 & 50.83 & 21.35 & 29.97\\
    Gemini-2.5-pro & 12.22 & 15.41 & 39.85 & 41.76 & 54.50 & 21.91 & 30.94\\
    \rowcolor{red!5} Gemini-3-pro & 26.37$^\spadesuit $ & 23.61 & 41.35 & 65.88$^\spadesuit $ & 58.94 & 26.97 & 40.52\\
    Qwen3-VL-8B-Inst. & 1.29 & 4.59 & 20.05 & 12.35 & 26.94 & 6.18 & 11.90\\
    Qwen3-VL-32B-Inst. & 2.25 & 4.59 & 24.81 & 17.65 & 24.81 & 11.24 & 14.22\\
    \midrule
    \multicolumn{8}{c}{\hspace{0.5cm}\textbf{Agent Workflow}}  \\
    \midrule
    GPT-5 & 31.61 & 52.13 & 51.63 & 77.65 & 72.28 & 25.84 & 51.86\\
    Gemini-2.5 Pro & 30.65 & 48.20 & 49.50 & 77.65 & 72.33 & 27.53 & 50.98\\
    Gemini-3 Pro & \extrabold{33.51} & 53.20 & \underline{51.78} & \extrabold{82.94} & \extrabold{76.67} & \underline{28.65} & \underline{54.46} \\
    \rowcolor{gray!15} Qwen3-VL-8B-Inst. & 18.38 & 32.13 & 35.89 & 57.06 & 64.22 & 11.80 & 36.58\\
    \rowcolor{gray!15} Qwen3-VL-32B-Inst. & 14.84 & 38.69 & 38.69 & 68.52 & 66.94 & 17.42 & 40.85\\
    \midrule
    \multicolumn{8}{c}{\textbf{\hspace{1cm}Multimodal Deep Search Agents}}  \\
    \midrule
    MMSearch-R1-7B & - & 20.33$^\dag $ & - & 53.80$^\dag $ & 58.40$^\dag $ & - & -\\
    SenseNova-MARS-8B & - & 41.64$^\dag $ & - & 67.84$^\dag $ & 67.11$^\dag $ & - & -\\
    SenseNova-MARS-32B & - & \extrabold{54.43}$^\dag $ & - & 74.27$^\dag $ & 72.61$^\dag $ & - & -\\
    Webwatcher-7B & - & - & 20.30$^\dag $ & 49.10$^\dag $ & - & - & -\\
    Webwatcher-32B & - & - & 26.70$^\dag $ & 55.30$^\dag $ & - & - & -\\
    MM-DeepResearch 32B & - & - & 43.00$^\dag $ & 69.00$^\dag $ & 70.10$^\dag $ & - & -\\
    \midrule
    \multicolumn{8}{c}{\hspace{0.5cm}\textbf{Ours}}  \\
    \midrule
    % DAPO-7B-replay & 27.62 & 43.61 & 43.86 & 76.47 & 71.33 & 15.17 & 46.34\\
\modelnamesmall & 26.77 & 47.54 & 44.36 & 79.41 & 73.06 & 20.79 & 48.66~\scriptsize{(+12.08)}\\
\modelnamebig & \underline{31.93} & \underline{53.95} & \extrabold{53.77} & \underline{82.35} & \underline{76.00} & \extrabold{29.78} & \extrabold{54.63}~\scriptsize{(+13.78)}\\
    \bottomrule
    \end{tabular}
    \end{threeparttable}
    \vspace{-1mm}
\caption{Accuracy (\%) on deep-search benchmarks. Best results are in \textbf{bold}, second-best are \underline{underlined}, $\dag$ denotes results reported in the original papers, and $\spadesuit $ indicates potential dataset contamination.}
    \label{tab:mmsearch_results}
    \vspace{-3mm}
    \end{table*}

\textbf{Potential Dataset Contamination in Model Training.}
Some datasets may have been inadvertently included in the training data of certain models. 
For example, MM-Search appears to be present in the training corpus of Gemini-3-Pro: while Gemini-2.5-Pro achieves $39.8\%$ on this benchmark, Gemini-3-Pro reaches $65.88\%$ even without the use of external search tools. 
This discrepancy suggests that performance gains may partially stem from prior exposure rather than improved reasoning or retrieval capabilities. Consequently, the reported results of such commercial LLMs on these benchmarks may not accurately reflect their true deep search abilities.

\textbf{Training Improves Agent Performance and Scales Effectively.} Table~\ref{tab:mmsearch_results}~demonstrates that training on \datasetname~significantly improves the performance of the base models. Compared to models without training but equipped with same tools (\S\ref{sec:agent_setup}), \modelnamesmall~achieves an average improvement of $12.08\%$, while \modelnamebig~achieves a $13.78\%$ gain in benchmark accuracy.
Performance scales consistently with model size: the 32B model outperforms the 8B variant both before and after training, while RL training yields complementary gains across scales, demonstrating the scalability of our approach.

\textbf{Competitive Performance Against State-of-the-Art Search Agents.}
Compared to specialized multimodal deep search agents (see the Multimodal Deep Search Agents rows in Table~\ref{tab:mmsearch_results}), our approach achieves competitive or superior results without requiring task-specific architectures, highlighting the generality of our training paradigm. 
Furthermore, when compared to state-of-the-art commercial LLMs under the same tool setting, our fine-tuned model \modelnamebig~also achieves comparable or even superior performance. Specifically, it outperforms GPT-5 by $2.78\%$ and Gemini-2.5-pro by $3.65\%$. Even when compared to Gemini-3-pro, which may benefit from potential training data contamination, our model achieves comparable performance ($54.63\%$ vs.\ $54.46\%$).

% Overall, these results show that RL training on our curated data effectively enhances both reasoning and tool-use capabilities, enabling open-source VLLMs to close the gap with leading commercial systems on challenging multimodal deep search benchmarks.

\vspace{-3mm}
\section{Analysis}
\label{sec:analysis}
\vspace{-1mm}
\subsection{Effect of the Data choice}
\label{sec:data_chocie}
\vspace{-1mm}

\begin{table}[]
% \footnotesize
\small
\centering
\setlength\tabcolsep{6pt}
\begin{tabular}{l|cccc|c}
Data Choice                    & MMSrch-Plus & BC-VL & MMSrch & HR-MMSrch & Avg. \\
\toprule
\rowcolor{gray!15} \textit{Qwen3-VL-8B-Inst.} & 18.38 & 35.89 & 57.06 & 32.13 & 35.87 \\
\quad \textit{w/} F+L & 20.96 & 43.22 & 73.46 & 43.28 & 45.23 \\
\quad \textit{w/} F + L + 4-hops (F+L) & 20.32 & 43.35 & 74.12 & 43.60 & 45.35 \\
\quad \textit{w/} F + L + 2/3/4 hops (F + L) & 20.96 & 44.11 & 75.29 & 44.61 & 46.24 \\
\quad \textit{w/} F + L + 2/3/4 hops (Full) & \extrabold{26.77} & \extrabold{44.36} & \extrabold{79.41} & \extrabold{47.54} & \extrabold{49.52}
\\ \bottomrule
\end{tabular}
\vspace{-1mm}
\caption{Impact of different training data choices for Qwen3-VL-8B-Inst. as the search agent backbone, using the same settings (100 steps, batch size = 64, rollout = 8).}
\label{tab:datachoices}
\vspace{-4mm}
\end{table}

We first train the search agent backbone using the FVQA and LiveVQA-News datasets (F+L). 
Table~\ref{tab:datachoices} shows that this setting improves performance over the Qwen3-VL-8B-Inst.~\citep{qwen3technicalreport} baseline without training, increasing the average score from $35.87\%$ to $45.23\%$ across all benchmarks. 
However, these datasets mainly contain news-related images, which are relatively simple and lack diversity.
To increase task difficulty, we extend the training data with generated 4-hop questions (F+L+4-hops (F+L)). 
This leads to only limited improvement, with the average score increasing slightly from $45.23\%$ to $45.35\%$, suggesting that these 4-hop questions are too challenging when used alone.
To address this, we further include intermediate 2-hop and 3-hop questions together with 4-hop questions (F+L+2/3/4 hops). 
As shown in the table, this results in consistent performance gains, improving the average score to $46.24\%$.
Finally, the full data mixture (F+L+2/3/4 hops (Full)), which incorporates InfoVQA, Infoseek, and OK-VQA, achieves the best results across all benchmarks, reaching an average of $49.52\%$. 
This highlights the importance of both diverse seed VQA data and a balanced range of question difficulties for training an effective search agent.

\vspace{-2mm}
\subsection{Analysis of Tool Usage}
\vspace{-1mm}

\begin{figure}[b!]
\centering
\begin{minipage}{0.45\textwidth}
    \centering
    \vspace{-3mm}
\includegraphics[width=\linewidth]{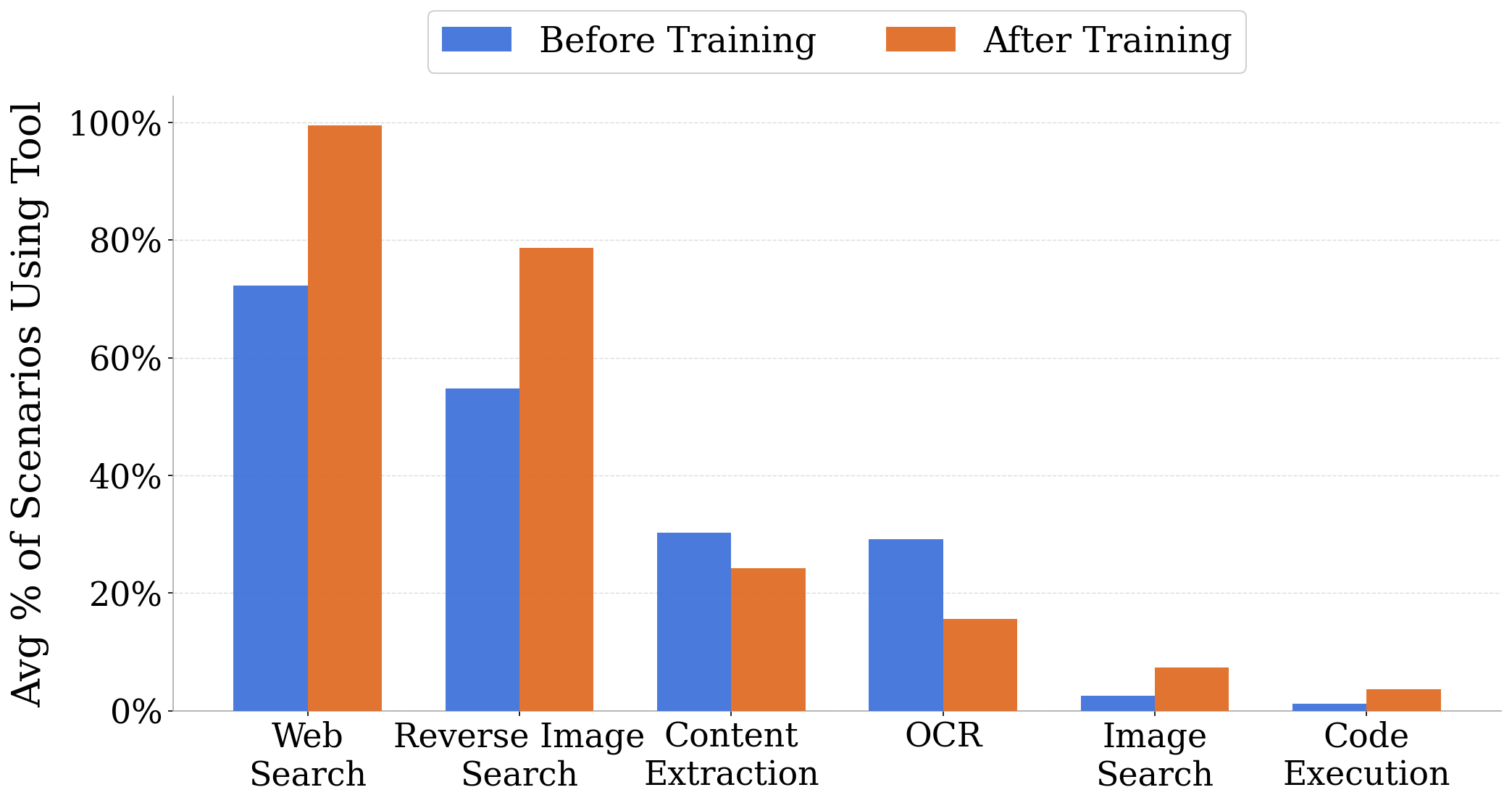}
\vspace{-5mm}
    \caption{Tool usage frequency across scenarios on all test datasets. Each bar shows the percentage of scenarios in which a given tool is used at least once.}
    \label{fig:tool}
\end{minipage}
\hfill
\begin{minipage}{0.45\textwidth}
    \centering
    % \vspace{2mm}
    \vspace{-3mm}
    \includegraphics[width=\linewidth]{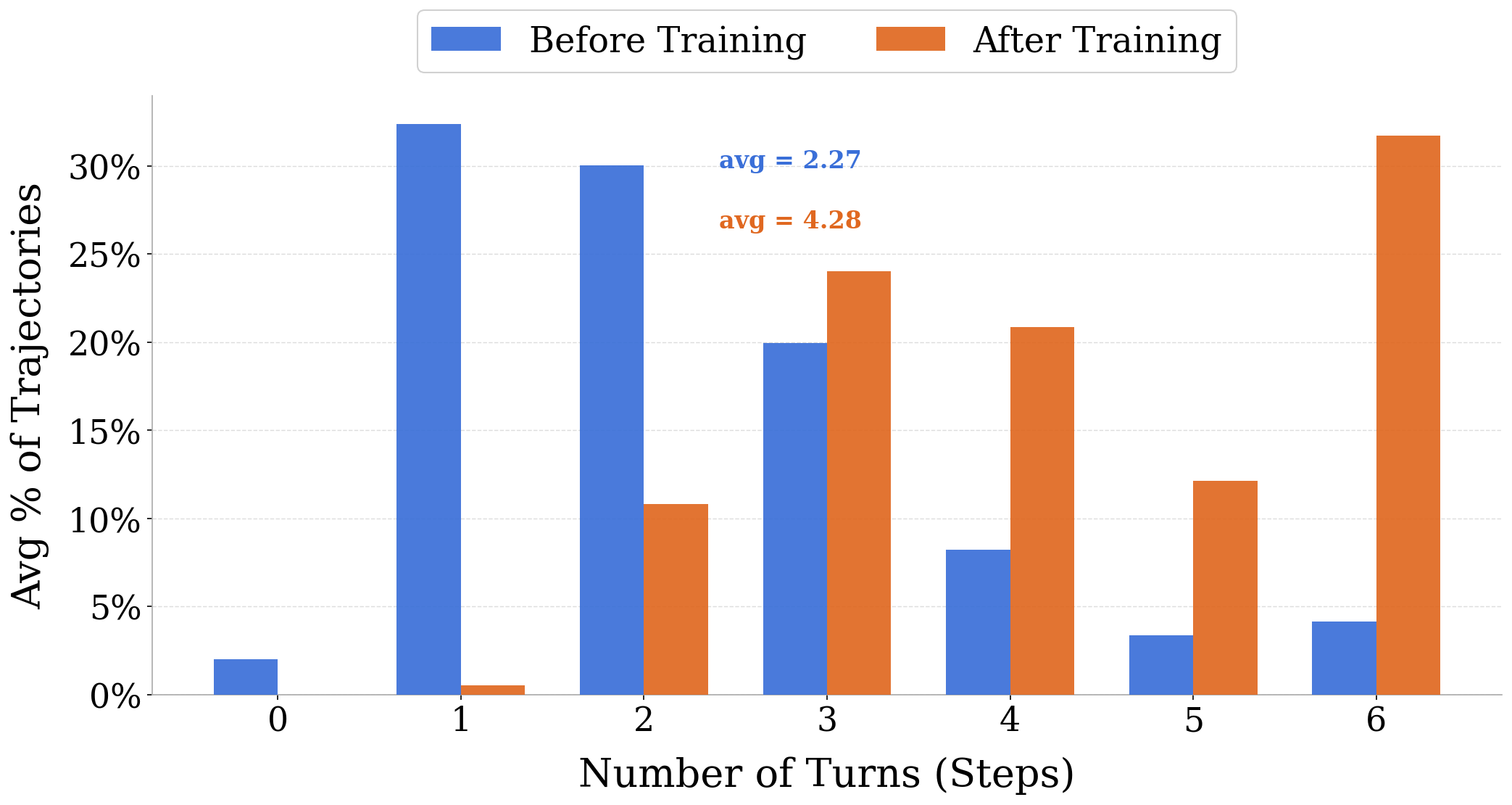}
    \vspace{-5mm}
    \caption{Turn distribution across scenarios on all test datasets. Each bar shows the percentage of trajectories that terminate at a given number of turns. }
    \label{fig:turn}
\end{minipage}
\end{figure}

We analyze how RL training reshapes both search depth (number of turns) and tool-use strategy, compared to the pre-trained model.

\textbf{Tool use.}
As shown in Figure~\ref{fig:tool},
before training, the model uses Web Search ($72\%$) and Reverse Image Search ($55\%$) at moderate rates with no clear retrieval strategy, indicating heuristic and unstructured tool selection. After RL training, tool use becomes highly structured: Web Search is nearly universal ($99\%$) and Reverse Image Search rises to $79\%$, forming a consistent two-stage retrieval strategy (text search followed by visual grounding). Content Extraction decreases to $24\%$ and OCR drops to $16\%$, suggesting a shift away from low-level image parsing toward active web-based retrieval.

\textbf{Search depth.}
As shown in Figure~\ref{fig:turn}, before training, the model exhibits a strongly left-skewed turn distribution: $32\%$ of trajectories stop after one turn and $30\%$ after two, with an average of $2.27$ turns, indicating shallow and non-persistent search. After RL training, the distribution shifts substantially rightward, with the largest share of trajectories reaching the 6-turn limit ($32\%$), followed by peaks at 3 ($24\%$) and 4 turns ($21\%$). The average increases to $4.28$, demonstrating sustained multi-step reasoning across all benchmarks.
% GPT-5 averages $3.15$ turns and shows a bimodal distribution, with concentrations at both 1 turn ($23\%$) and 6 turns ($21\%$), suggesting it either resolves questions quickly from internal knowledge or requires extended retrieval when it cannot---rarely falling in between.

The pre-trained multimodal language model behaves as a shallow agent with inconsistent tool use. RL training transforms it into a systematic multi-step search agent with longer trajectories and a consistent retrieval pipeline. 
% Compared to GPT-5, the trained model adopts a more aggressive search strategy, emphasizing exploration and evidence gathering, whereas GPT-5 more often relies on its internal knowledge to reduce or bypass search.
All results are in Appendix~\ref{app:tool_turn_analysis}.

\vspace{-2mm}
\subsection{Training without External Tool Calls}
\vspace{-1mm}
\label{sec:cache_train}

\begin{table*}[t]
\centering
% \footnotesize
\small
\centering
\setlength\tabcolsep{3.2pt}
\begin{threeparttable}
\begin{tabular}{l|cccccc|c}
\toprule
Model
& MMSrch-Plus
& HR-MMSrch 
& BC-VL 
& MMSrch
& FVQA
& MTA-test
& Avg\\
% \cmidrule(lr){2-3}
% & Easy & Hard & Test & Test & Test & Test \\
\midrule
\rowcolor{gray!15} \textit{Qwen3-VL-8B-Inst.}
& 18.38
& 32.13
& 35.89    
& 57.06  
& 64.22
& 11.80 
& 36.58 \\
\quad + FVQA \& LiveVQA & 20.96 & 43.28 & 43.22 & 73.46 & 71.06
& 12.36 & 44.06\\
\rowcolor{blue!10} \quad + MTA-DS (replay)
& 27.62    
& 43.61 
& 43.86 
& 76.47 
& 71.33
& 15.17
& 46.34 \\
\quad + MTA-DS (tool-call)    
& 26.77
& 47.54 
& 44.36 
& 79.41 
& 73.06 
& 20.79 
& 48.66 \\
\bottomrule
\end{tabular}
\end{threeparttable}
\vspace{-1mm}
\caption{Accuracy (\%) on search-oriented benchmarks with and without external tool calls.}
\vspace{-3mm}
\label{tab:replay}
\end{table*}

Training a search agent is expensive, as it requires not only GPUs but also frequent tool usage (e.g., paid web search APIs). To reduce this cost, we explore whether the data generated during QA creation can be reused for RL training.
We store all rollout histories from the data creation pipeline and ablation studies in a replay dictionary, where keys correspond to tool inputs (parameters) and values correspond to tool outputs. During training, instead of calling external tools, we retrieve responses from this dictionary. Specifically, we compute the cosine similarity between the current tool query and stored keys, and select the most similar entry if the similarity exceeds $0.75$. If no match is found, a tool error is returned.
As shown in Table~\ref{tab:replay}, this replay-based approach outperforms the baseline and even achieves better performance than training with FVQA and LiveVQA data using real tool calls. This result suggests that, with our constructed dataset, it is possible to train an effective search agent without incurring additional tool costs, enabling efficient training of open-source models. More details about how we create the cache are shown in Appendix~\ref{app:cache}.
% \jw{TODO}.

\vspace{-2mm}
\section{Conclusion}
\vspace{-1mm}

In this work, we explore how to address a key limitation of Multimodal Large Language Models (MLLMs): their difficulty in complex, multi-step visual reasoning that requires deep search and evidence integration. 
We propose \framename, a synthesis framework that generates high-quality multi-hop vision-language trajectories. Using a tool-augmented agent over existing VQA datasets, we construct \datasetname, a 21K-example dataset designed for long-horizon reasoning.
RL training on this data significantly improves open-source models. Our \modelnamebig~achieves state-of-the-art performance across six benchmarks, outperforming GPT-5 and Gemini-3-Pro under comparable tool settings. It also improves search behavior, increasing average search depth from $1.76$ to $3.62$ turns and enabling more systematic multi-step retrieval. We further show that training via interaction replay is effective, reducing the cost of tool usage.
Our results highlight three key insights: (1) high-quality multi-hop data enables strong long-horizon reasoning; (2) data diversity and verification are critical for performance; and (3) replay-based training improves efficiency without sacrificing accuracy. By releasing our open recipe, dataset, test suite, and rollout histories, we provide a foundation for future multimodal deep-search agents.

\clearpage

\bibliography{colm2026_conference}
\bibliographystyle{colm2026_conference}

\appendix
\newpage

\section*{LLM Disclosure}
We used large language models as tools throughout the data synthesis, verification, and evaluation pipeline. In particular, GPT-5-mini was used to filter seed VQA samples for image dependence; GPT-5 and GPT-5.1 were used to rewrite questions, generate candidate hop-level QA pairs, and disambiguate entity answers; and GPT-5 with web search was used as an independent verifier of factual consistency, answer uniqueness, and temporal stability. We also used an LLM-based judge to compute training rewards and evaluate rollout quality during reinforcement learning. In addition, GPT-5, Gemini-2.5-Pro, and Gemini-3-Pro were used as baseline systems in our experiments.

\newpage
\section{Method Details}

\subsection{Definition of an Entity Answer}
\label{app:entity_def}

A core requirement of our dataset is that the \emph{seed answer}---the answer to the first hop, which must be visually grounded---is a \textbf{named proper-noun entity}: a specific, uniquely identifiable real-world referent rather than a generic descriptor, number, or action. Concretely, we define an entity as any answer belonging to one of the following ten categories:

\begin{enumerate}[leftmargin=1.5em, itemsep=2pt]
    \item \textbf{People.} Famous individuals who are uniquely named, including actors, musicians, scientists, politicians, historical figures, and athletes (e.g., \textit{Marie Curie}, \textit{Lionel Messi}).
    \item \textbf{Organizations.} Named institutions such as technology companies, NGOs, universities, sports teams, and government bodies (e.g., \textit{UNICEF}, \textit{Manchester United}).
    \item \textbf{Locations.} Specific geographic or architectural referents, including cities, countries, landmarks, airports, and nature reserves (e.g., \textit{Eiffel Tower}, \textit{Yellowstone}).
    \item \textbf{Products.} Commercially named goods such as consumer electronics, vehicles, and software (e.g., \textit{iPhone~15}, \textit{Tesla Model~3}).
    \item \textbf{Events.} Named occurrences with a defined scope, including global sporting events, historical events, conferences, and cultural festivals (e.g., \textit{FIFA World Cup}, \textit{Diwali}).
    \item \textbf{Creative Works.} Titled artistic or intellectual productions, including films, television series, books, artworks, and music albums (e.g., \textit{Inception}, \textit{Mona Lisa}).
    \item \textbf{Science \& Technology.} Named scientific entities such as biological species, chemical compounds, devices, AI models, and benchmark datasets (e.g., \textit{GPT-4}, \textit{ImageNet}).
    \item \textbf{Medical \& Health.} Named diseases or medical conditions (e.g., \textit{COVID-19}, \textit{Malaria}).
    \item \textbf{Financial \& Business.} Named companies, stock indexes, and commercial brands (e.g., \textit{Coca-Cola}, \textit{S\&P~500}).
    \item \textbf{Space \& Astronomy.} Named celestial objects, space missions, and astronomical events (e.g., \textit{Hubble Telescope}, \textit{Mars}).
\end{enumerate}

\noindent
Answers that are numbers, adjectives, generic nouns (e.g., ``a bridge'', ``the car''), actions, or multi-entity phrases are explicitly excluded. This constraint ensures that every seed answer is (i)~factually verifiable through web search, (ii)~a well-defined node in a real-world knowledge graph, and (iii)~unambiguously linkable to follow-on questions in the multi-hop chain.

\newpage
\subsection{VQA Data Filtering Pipeline}
\label{app:seed}

We describe the filtering pipeline used to extract high-quality entity-grounded VQA pairs from raw datasets. 
Each candidate question-answer pair passes through five sequential stages; a pair is accepted only if it passes every stage. 

\subsubsection*{Stage 1: Vision Requirement Check}

We first verify that the question \emph{genuinely requires} visual information to answer. Questions answerable from general knowledge alone (e.g., ``What is the capital of France?'') are discarded. This check is performed by GPT-5-mini with the following prompt:

\begin{tcolorbox}[
  colback=gray!8, colframe=gray!50, 
  title=\textbf{Prompt: Vision Requirement Check},
  fonttitle=\small\bfseries, fontlower=\small,
  ]
\small
\texttt{Analyze this question and determine if it requires visual information from an image to answer:}

\texttt{Question: \{question\}}

\medskip
\texttt{Consider these scenarios that REQUIRE an image to provide a UNIQUE answer:}
\begin{itemize}[noitemsep, topsep=2pt, leftmargin=1.2em]
  \item Questions about specific visual elements that make the answer unique (``What team do the standing players play for?'')
  \item Questions about text, signs, names, or writing visible in the image (``What name can be found on the blackboard?'')
  \item Questions about specific people, objects, or entities shown in the image (``Who is this person?'', ``What company logo is shown?'')
  \item Questions about visual attributes that identify specific entities (``What color jersey is the team wearing?'')
  \item Questions about spatial relationships that identify specific locations (``Where are these animals located?'')
  \item Questions using demonstrative pronouns referring to specific visual elements (``this person'', ``that building'', ``these objects'')
  \item Questions where the image provides the \emph{only} way to determine the specific answer (not general knowledge)
\end{itemize}

\medskip
\texttt{Consider these scenarios that DO NOT require an image:}
\begin{itemize}[noitemsep, topsep=2pt, leftmargin=1.2em]
  \item General knowledge questions with standard answers (``What is the capital of France?'')
  \item Questions about well-known facts that don't depend on visual context (``When was Einstein born?'')
  \item Questions about definitions, explanations, or concepts that are universally known
  \item Questions where the answer can be determined from text alone
  \item Questions that would have the same answer regardless of what image is shown
\end{itemize}

\medskip
\texttt{Please only return `Yes' if the question REQUIRES an image to answer, `No' if it can be answered without visual information.}
\end{tcolorbox}

\subsubsection*{Stage 2: Q\&A Rewriting and Free-Form Validity Check}

Raw VQA answers often contain chain-of-thought reasoning traces, and questions may be phrased as multiple-choice. We use GPT-5 to simultaneously (i)~rewrite the question into a clean free-form format that explicitly references ``the image'', (ii)~extract the concise answer entity from any reasoning trace, (iii)~produce a \texttt{complete\_answer} with minimal disambiguating context (at most three words), and (iv)~flag questions that are \emph{inherently multiple-choice}---i.e., questions that cannot be rephrased as free-form without providing specific comparison options (e.g., ``This person is playing a similar sport to whom?''). Such questions are rejected at this stage.

\begin{tcolorbox}[
  colback=gray!8, colframe=gray!50,
  title=\textbf{Prompt: Q\&A Rewriting},
  fonttitle=\small\bfseries,
  ]
\small
\texttt{Here is a question and answer pair where the answer contains reasoning:}

\texttt{Please rewrite this as a clean question-answer pair where:}
\begin{enumerate}[noitemsep, topsep=2pt, leftmargin=1.5em]
  \item The question should mention ``in the image'' and remove any answer choices.
  \item Ensure the question makes sense as a free-form question without choices. Questions like ``This person is playing a similar sport to whom?'' require choices to be answerable and should be marked as invalid.
  \item If the question cannot be reasonably converted to free-form without choices, respond with \texttt{\{``valid'': false, ``reason'': ``Question requires multiple choice options to be answerable''\}}.
  \item The answer should be just the final entity/answer without any reasoning or explanation.
  \item Provide both a short answer (\texttt{output\_answer}) and a \texttt{complete\_answer}:
    \begin{itemize}[noitemsep, leftmargin=1.2em]
      \item Add descriptive context \emph{from the provided context} to disambiguate it.
      \item Keep the new description less than 3 words---add the minimum context needed for disambiguation.
    \end{itemize}
\end{enumerate}

\medskip
\textit{Examples of disambiguation:}
\begin{itemize}[noitemsep, topsep=2pt, leftmargin=1.2em]
  \item ``Apple'' (technology company) $\to$ \texttt{complete\_answer}: ``Apple, a technology company''
  \item ``Birmingham'' (city in Alabama) $\to$ \texttt{complete\_answer}: ``Birmingham, city in Alabama''
  \item ``Baltimore Orioles'' (baseball team) $\to$ \texttt{complete\_answer}: ``Baltimore Orioles, a Baseball team''
\end{itemize}

\medskip
\texttt{Format your response as JSON:}
\begin{verbatim}
// If valid:
{
  "valid": true,
  "output_question": "[clean question mentioning 'in the image']",
  "output_answer": "[direct short answer only]",
  "complete_answer": "[add context ONLY if ambiguous AND
                       ONLY using info from input_answer]"
}
// If invalid:
{
  "valid": false,
  "reason": "[explanation why it cannot be converted]"
}
\end{verbatim}

\texttt{Input:}
\begin{verbatim}
{
  "input_question": "{question}",
  "input_answer": "{answer}"
}
\end{verbatim}
\end{tcolorbox}

\subsubsection*{Stage 3: Named Entity Classification}

We verify that the rewritten answer is a \emph{named proper noun entity} belonging to one of ten high-level categories (Table~\ref{tab:entity_categories}). This check is performed by GPT-5. Answers that are numbers, generic terms, actions, or vague phrases are rejected.

\begin{table}[h]
\centering
\small
\begin{tabular}{ll}
\toprule
\textbf{Category} & \textbf{Example entities} \\
\midrule
People & Barack Obama, Marie Curie, Lionel Messi \\
Organizations & Google, UNICEF, Manchester United \\
Locations & Tokyo, Eiffel Tower, Yellowstone \\
Products & iPhone 15, Tesla Model 3, Photoshop \\
Events & FIFA World Cup, Diwali, TED Conference \\
Creative Works & Inception, Mona Lisa, Harry Potter \\
Science \& Technology & GPT-4, ImageNet, Bengal tiger \\
Medical \& Health & COVID-19, Malaria \\
Financial \& Business & Coca-Cola, S\&P 500 \\
Space \& Astronomy & Mars, Hubble Telescope \\
\bottomrule
\end{tabular}
\caption{Entity categories used for filtering.}
\label{tab:entity_categories}
\end{table}

\begin{tcolorbox}[
  colback=gray!8, colframe=gray!50,
  title=\textbf{Prompt: Named Entity Classification},
  fonttitle=\small\bfseries,
  ]
\small
\texttt{Here is a question and answer pair:}\\
\texttt{Question: \{question\}}\\
\texttt{Answer: \{answer\}}

\medskip
\texttt{Check carefully if the answer is a named, proper noun entity from these specific categories:}

\textit{[Full category list as in Table~\ref{tab:entity_categories}, with subcategory examples]}

\medskip
\texttt{Note: The reasoning process is not part of the answer! Answers with multiple entities or non-entity answers should be returned as `No'. Exclude: Numbers, actions, generic entities, or vague terms.}

\medskip
\texttt{Please only return `Yes' or `No'.}
\end{tcolorbox}

\subsubsection*{Stage 4: Entity Validity Verification}

As a secondary guard, GPT-5-mini independently checks whether the extracted answer string constitutes a valid, specific proper noun (e.g., rejecting generic terms that may have slipped through Stage~3). The prompt mirrors the category taxonomy in Table~\ref{tab:entity_categories} and again requests a binary \texttt{Yes}/\texttt{No} response.

\begin{tcolorbox}[
  colback=gray!8, colframe=gray!50,
  title=\textbf{Prompt: Entity Validity Verification},
  fonttitle=\small\bfseries,
  ]
\small
\texttt{Is this phrase a valid, proper noun entity? \{entity\}.}

\medskip
\texttt{I am looking for specific categories of entities:}

\textit{[Full category list with subcategory examples, same as Stage 3]}

\medskip
\texttt{Exclude: Numbers, actions, generic entities, or vague terms.}

\medskip
\texttt{Please only return `Yes' or `No'.}
\end{tcolorbox}

\subsubsection*{Stage 5: Visual Answer Verification}

Finally, GPT-5 with vision and web search is used to confirm that the proposed answer is factually correct given the image(s). This guards against errors introduced by the original dataset or the rewriting step. The model receives the image(s) via the Responses API and is permitted to issue web search queries to verify entity-specific facts. A pair is accepted only if the model responds with a definitive \texttt{Yes}; an \texttt{Unsure} response triggers rejection.

\begin{tcolorbox}[
  colback=gray!8, colframe=gray!50,
  title=\textbf{Prompt: Visual Answer Verification (GPT-5 with Vision + Web Search)},
  fonttitle=\small\bfseries,
  ]
\small
\texttt{You are an expert at verifying answers to visual questions. You have extensive knowledge about entities, landmarks, people, brands, and locations worldwide.}

\medskip
\texttt{Question: \{question\}}\\
\texttt{Proposed Answer: \{answer\}}

\medskip
\texttt{Please carefully analyze the image(s) and determine if the proposed answer is correct. Use your extensive knowledge and web search to verify factual claims about entities.}

\medskip
\texttt{Important:}
\begin{itemize}[noitemsep, topsep=2pt, leftmargin=1.2em]
  \item The answer should be factually accurate based on what is visible in the image(s).
  \item Consider the answer correct if it is semantically equivalent (e.g., ``NYC'' and ``New York City'').
  \item Be strict about factual accuracy---if the answer is wrong, provide the correct answer.
  \item Use your knowledge of world facts, famous people, landmarks, brands, and locations to verify.
  \item Use web search when needed to verify specific facts.
\end{itemize}

\medskip
\texttt{Respond with ONE of the following:}
\begin{itemize}[noitemsep, topsep=2pt, leftmargin=1.2em]
  \item \texttt{"Yes"} if the answer is correct.
  \item \texttt{"Unsure"} if you cannot determine correctness with confidence.
\end{itemize}
\end{tcolorbox}

\subsubsection*{Pipeline Summary}

Table~\ref{tab:filter_summary} summarizes the five stages, the model used at each stage, and the rejection criterion.

\begin{table}[h]
\centering
\small
\caption{Summary of the VQA filtering pipeline.}
\label{tab:filter_summary}
\begin{tabular}{clll}
\toprule
\textbf{Stage} & \textbf{Check} & \textbf{Model} & \textbf{Reject if\ldots} \\
\midrule
1 & Vision requirement & GPT-5-mini & Question answerable without image \\
2 & Q\&A rewriting & GPT-5 & Question cannot be free-form; conversion fails \\
3 & Entity classification & GPT-5 & Answer is not a named proper noun entity \\
4 & Entity validity & GPT-5-mini & Answer is a generic or vague term \\
5 & Visual verification & GPT-5 (vision + web) & Answer is factually incorrect or uncertain \\
\bottomrule
\end{tabular}
\end{table}

\subsection{Tools}
\label{app:tools}

The QA agent is equipped with four tools that cover complementary modalities of information retrieval:

\begin{enumerate}[leftmargin=1.5em, itemsep=4pt]

    \item \textbf{Web Search.}
    Given a text query, this tool issues a search request via the Tavily API and returns up to 10 results, each consisting of a URL, title, short snippet, and---when available---the full extracted page content in Markdown format. Queries are constrained to always include the complete disambiguated entity name to prevent irrelevant retrievals.

    \item \textbf{Web Reader.}
    Given a URL, this tool fetches and returns the full textual content of the target webpage. It is used to deepen coverage of a specific source when the snippet returned by web search is insufficient for synthesizing a well-grounded question.

    \item \textbf{Google Lens.}
    Given an image, this tool performs a reverse image search and returns visually similar results with their associated titles and snippets. It is particularly useful for identifying specific objects, landmarks, or entities that are visually distinctive but difficult to describe in text alone.

    \item \textbf{Image Search.}
    Given a text query, this tool retrieves image URLs along with auto-generated descriptions and captions. It is used when visual context about the current entity---such as product appearance, architectural style, or logo design---can surface facts that text-only pages do not explicitly state.

\end{enumerate}

\noindent
Together, these tools allow the agent to gather evidence through both text and visual channels. At each hop, the agent selects the most appropriate tool and query based on the nature of the current entity and the type of bridging fact it seeks to uncover.

\subsection{Context Chunking}
\label{app:chunking}

\paragraph{Source Selection.}
GPT-5 selects up to three sources from the retrieved results by scoring each
candidate on three criteria: (1) the content must mention $a_{k-1}$; (2) it
should contain non-obvious facts unlikely to be known without retrieval; and
(3) it must include proper-noun entities that could serve as the next answer
$a_k$. Previously visited URLs are excluded to encourage diversity across hops.

\paragraph{Passage Selection.}
Each selected source is split into fixed-length windows. Each window is scored
by a lightweight heuristic: it receives $+10$ per complete sentence and $+2$
per occurrence of common English function words (e.g., \textit{the},
\textit{and}, \textit{is}), rewarding coherent natural-language prose over
boilerplate or markup; windows containing HTML or JSON structures are
penalized. Windows that contain $a_{k-1}$ receive a large additive bonus to
prioritize entity-relevant passages. The top-scoring windows are concatenated
in reading order to form the context passed to the Q\&A extraction step.

\subsection{Q\&A Candidate Generation}
\label{app:qa_synthesis}

For each retrieved source, the system runs a four-step pipeline to produce a verified Q\&A candidate $(q, a, c)$, where $q$ is the question, $a$ is the answer entity, and $c$ is the supporting context excerpt. One candidate is attempted per source; across up to ten sources per hop, this yields a pool of candidates that are later merged and validated.

\subsubsection{Step 1: Content Summarization.}
Raw webpage content is often long and only partially relevant. GPT-5 first produces a focused 200--400 word summary of the page chunk, emphasizing facts about the current entity that could serve as Q\&A fodder---specific dates, locations, relationships, affiliated organizations, and other named entities. Generic well-known facts are explicitly excluded. The summarization prompt is:

\begin{tcolorbox}[colback=gray!8, colframe=gray!50,
  title=\textbf{Prompt: Content Summarization (GPT-5)},
  fonttitle=\small\bfseries, ]
\small
\texttt{Summarize the key facts from this webpage content that are relevant to ``\{entity\_name\}''.}

\medskip
\texttt{Title: \{title\}\\
Search Query: \{search\_query\}\\
Content: \{raw\_content\}}

\medskip
\texttt{Focus on:}
\begin{itemize}[noitemsep, topsep=2pt, leftmargin=1.2em]
  \item Facts about \{entity\_name\} (dates, locations, relationships, achievements)
  \item Other entities mentioned (people, companies, places) that could be answer candidates
  \item Specific numbers, dates, or names (these make good answers)
\end{itemize}

\medskip
\texttt{Provide a concise summary (200--400 words) highlighting the most interesting and specific facts.
Do NOT include generic information that everyone knows.}
\end{tcolorbox}

\subsubsection{Step 2: Q\&A Extraction.}
\label{app:qa_gen}
Given the summarized content, GPT-5.1 extracts a single factual question--answer pair. For all hops except the final one, the answer must be a named proper-noun entity; for the final hop, numeric and date answers are also permitted to widen coverage. Six hard constraints are enforced: (1) the question must contain the current entity name; (2) the answer must be unique---exactly one correct answer; (3) the answer must not duplicate any previous hop's answer; (4) removing the entity from the question must make the answer indeterminate; (5) the question must be standalone, without references to ``the passage'' or ``the context''; and (6) the question must ask about a meaningful relationship or attribute.

\begin{tcolorbox}[colback=gray!8, colframe=gray!50,
  title=\textbf{Prompt: Q\&A Extraction (GPT-5.1, intermediate hops)},
  fonttitle=\small\bfseries, ]
\small
\texttt{You will be given content about ``\{entity\_name\}''. Find a simple factual question that involves ``\{entity\_name\}'' and has a specific entity as the answer.}

\medskip
\texttt{The answer should be a proper noun entity from one of these categories:\\
\{entity\_categories\}}

\medskip
\texttt{Search Query Used: \{search\_query\}\\
Content: \{content\}}

\medskip
\texttt{AVOID DUPLICATE ANSWERS:\\
Previous answers already used: \{previous\_answers\}\\
Your new answer MUST be different from ALL previous answers listed above.}

\medskip
\texttt{CRITICAL REQUIREMENTS:}
\begin{enumerate}[noitemsep, topsep=2pt, leftmargin=1.5em]
  \item The question must have a UNIQUE (singular) answer --- exactly ONE correct answer
  \item The question must include ``\{entity\_name\}'' in the question
  \item The question must be standalone (no references to ``the passage'' or ``the context'')
  \item Removing ``\{entity\_name\}'' from the question should make it have multiple possible answers
  \item The question should be USEFUL and NATURAL
  \item Ask about meaningful relationships, key facts, or important attributes
  \item Do not add any redundant details in the questions
\end{enumerate}

\medskip
\texttt{The question must have ONLY ONE POSSIBLE ANSWER --- AVOID:}
\begin{itemize}[noitemsep, topsep=2pt, leftmargin=1.2em]
  \item Multiple possible answers (e.g., ``What movies has X appeared in?'')
  \item Time-dependent answers without time constraint (e.g., ``Who is the current CEO?'')
  \item Subjective answers (e.g., ``What is the best...'')
  \item Weak relational phrases (``associated with'', ``related to'', ``connected to'')
  \item References to source material (``in the contexts'', ``mentioned in'')
\end{itemize}

\medskip
\texttt{Return a JSON object:}
\begin{verbatim}
{
  "question": "<question with {entity_name},
               ONLY ONE possible answer>",
  "answer":   "<specific entity name>",
  "contexts": "<exact text excerpt supporting this Q&A>"
}
\end{verbatim}
\texttt{If no suitable question can be created, return:\\
\{"question": "", "answer": "", "contexts": ""\}}
\end{tcolorbox}

\subsubsection{Step 3: Question Simplification.}
Extracted questions sometimes contain redundant temporal or locational qualifiers that do not contribute to making the answer unique (e.g., ``Which company is the first company Elon Musk created in 1995?''\ where the year is unnecessary). GPT-5.1 checks for and removes such redundancies while guaranteeing that the simplified question retains the same unique answer and still contains the entity name.

\begin{tcolorbox}[colback=gray!8, colframe=gray!50,
  title=\textbf{Prompt: Question Simplification (GPT-5.1)},
  fonttitle=\small\bfseries, ]
\small
\texttt{You are given a question that may have redundant or unnecessary details. Simplify it while preserving the unique answer.}

\medskip
\texttt{Original Question: \{question\}\\
Answer: \{answer\}\\
Entity in Question: \{entity\_name\}}

\medskip
\texttt{Remove unnecessary constraints/details that don't contribute to making the answer unique.}

\medskip
\texttt{Guidelines:}
\begin{enumerate}[noitemsep, topsep=2pt, leftmargin=1.5em]
  \item Keep the question grammatically correct and natural
  \item The simplified question must still have the SAME unique answer
  \item Remove redundant temporal info (years, dates) if not needed for uniqueness
  \item Remove redundant location info if not needed for uniqueness
  \item Keep the entity ``\{entity\_name\}'' in the question
\end{enumerate}

\medskip
\texttt{Do NOT simplify if all details are necessary for uniqueness, or if the question is already concise.}

\medskip
\texttt{Return:}
\begin{verbatim}
{
  "needs_simplification": <true|false>,
  "reason": "<why simplification is/isn't needed>",
  "simplified_question": "<simplified question, or
                           original if no change needed>"
}
\end{verbatim}
\end{tcolorbox}

\subsubsection{Step 4: Factual Verification.}
\label{app:qa_verification}

The (simplified) question and answer are verified by GPT-5.1 against five criteria before being accepted as a candidate. A candidate is only accepted if it passes \emph{all} five checks; otherwise it is discarded and the next source is tried.

\begin{tcolorbox}[colback=gray!8, colframe=gray!50,
  title=\textbf{Prompt: Q\&A Verification (GPT-5.1)},
  fonttitle=\small\bfseries, ]
\small
\texttt{Verify if this question-answer pair is factually correct and has a unique answer:}

\medskip
\texttt{Question: \{question\}\\
Answer: \{answer\}\\
Entity: \{entity\_name\}\\
Source URL: \{url\}\\
Context: \{context\}}

\medskip
\texttt{Please verify:}
\begin{enumerate}[noitemsep, topsep=2pt, leftmargin=1.5em]
  \item Is the answer factually correct according to the given context?
  \item Does the question contain the given entity ``\{entity\_name\}''?
  \item Is this answer unique (not multiple possible answers)?
  \item Is this answer stable over time (not changing)?
  \item Removing ``\{entity\_name\}'' from the question, is it impossible to get the unique answer?
\end{enumerate}

\medskip
\texttt{Return ``VERIFIED'' if all criteria are met, ``REJECTED'' if any fail.\\
Explain your reasoning briefly.}
\end{tcolorbox}

\noindent
Candidates that pass verification receive a \emph{complete answer}---a short disambiguating phrase appended to the bare answer (e.g., \textit{``Birmingham, city in Alabama''})---generated by GPT-5.1 from the supporting context.

\subsection{Candidate Selection via Difficulty Filtering}
\label{app:candidate_selection}

When multiple Q\&A candidates are generated from different sources, we select the one that produces the most challenging merged question rather than the most
semantically diverse answer. 
For each candidate, we measure how easily a weak model could reconstruct the reasoning path. Concretely, Qwen3-VL-32B is prompted to produce the search query it would use to answer the merged question $\tilde{q}_k$. This weak-model query $r_\text{weak}$ is then compared against the actual query $r_\text{actual}$ used to retrieve the supporting evidence,
using token-overlap Jaccard similarity after removing stop words:

\[
  \text{sim}(r_\text{weak},\, r_\text{actual})
  = \frac{|\mathcal{T}(r_\text{weak}) \cap \mathcal{T}(r_\text{actual})|}
         {|\mathcal{T}(r_\text{weak}) \cup \mathcal{T}(r_\text{actual})|}
\]

where $\mathcal{T}(\cdot)$ denotes the set of content tokens after stop-word removal. 
A low similarity score indicates that the weak model would search
differently from how the evidence was actually found, implying the question requires genuine multi-hop reasoning. 
We select the candidate with the lowest similarity score, subject to a threshold of $0.6$; if all candidates exceed
this threshold, the hop is rejected and a new search query is attempted.
Answer diversity across hops is enforced separately at the prompt level: the
Q\&A synthesis prompt explicitly lists all previous answers
$\{a_1, \ldots, a_{k-1}\}$ and instructs the model to generate an answer
distinct from all of them.

\subsection{Answer Disambiguation}
\label{app:disambiguation}

Short entity answers (e.g., \textit{Apple}) are often ambiguous and may cause
the next-hop search to retrieve irrelevant results. After a candidate
$(q_k, a_k, c_k)$ passes verification, GPT-5.1 appends a short disambiguating
phrase drawn from the supporting context $c_k$ to produce the complete answer
$\tilde{a}_k$. The disambiguated form is used exclusively as the seed for the
search query at hop $k{+}1$; the original $a_k$ is retained as the
ground-truth answer label.

\begin{tcolorbox}[colback=gray!8, colframe=gray!50,
  title=\textbf{Prompt: Answer Disambiguation (GPT-5.1)},
  fonttitle=\small\bfseries, ]
\small
\texttt{Given a question, answer, and supporting context, create a
``complete answer'' that adds descriptive information to make the answer
unique and unambiguous.}

\medskip
\texttt{Context: \{context\}\\
Question: \{question\}\\
Short Answer: \{answer\}}

\medskip
\texttt{RULES:}
\begin{enumerate}[noitemsep, topsep=2pt, leftmargin=1.5em]
  \item Add descriptive context \emph{from the provided context} to
        disambiguate the answer.
  \item Keep the added description to fewer than three words.
\end{enumerate}

\medskip
\textit{Examples:}
\begin{itemize}[noitemsep, topsep=2pt, leftmargin=1.2em]
  \item \texttt{``Apple''} $\to$ \texttt{``Apple, a technology company''}
  \item \texttt{``Birmingham''} $\to$ \texttt{``Birmingham, city in Alabama''}
  \item \texttt{``Baltimore Orioles''} $\to$
        \texttt{``Baltimore Orioles, a Baseball team''}
  \item \texttt{``google.com''} $\to$ \texttt{``google.com, a website''}
  \item \texttt{``Comcast''} $\to$ \texttt{``Comcast, a company''}
\end{itemize}

\medskip
\texttt{Return ONLY the complete answer, nothing else.}
\end{tcolorbox}

\noindent
The disambiguating phrase is always grounded in $c_k$ and is never
hallucinated. The length constraint (fewer than three words) ensures the
phrase remains concise enough to serve as an effective search seed without
over-specifying the query.

\subsection{Tool Response Cache Construction}
\label{app:cache}

To accelerate training and avoid redundant API calls during reinforcement learning rollouts, we build a static tool-response cache that maps structured lookup keys to pre-collected tool outputs. The cache covers all five tools used by the agent: \textbf{web search}, \textbf{web read}, \textbf{image search}, \textbf{reverse image search}, and \textbf{OCR}. Below we describe the two data sources, the key-construction scheme, and the quality filtering applied to each tool.

% ------------------------------------------------------------------
\subsubsection{Data Sources}

The cache is populated from two complementary sources.

\paragraph{Rollout JSONL files (training-time).}
During early training iterations, the agent's rollout records are saved as JSONL files. Each record contains a \texttt{tool\_interact\_info} list, where each element stores the raw action string—with tool-specific XML tags such as \texttt{<text\_search\_text>}, \texttt{<web\_read>}, \texttt{<ocr\_tool>}, \texttt{<text\_search\_image>}, and \texttt{<image\_search\_text>}—and the corresponding observation list. Extraction scripts parse these files, recover the tool-invocation parameters and raw responses, and emit a flat \texttt{\{cache\_key $\to$ response\}} JSON dictionary.

\paragraph{Trajectory JSON files (inference-time).}
We also save full trajectories as per-question JSON files during training data creation process. Each trajectory encodes the question context and the sequence of action steps, each with both a raw \texttt{observation} and an LLM-generated \texttt{observation\_summary}. A second set of extraction scripts parses these files and builds two parallel dictionaries: one mapping context-aware keys to observation summaries, and one mapping keys to raw observations.

% ------------------------------------------------------------------
\subsubsection{Cache Key Construction}
\label{appendix:cache:keys}

A central design choice is that cache keys are \emph{context-sensitive}: the same tool invocation may legitimately return a differently framed response depending on the question being answered. Keys are therefore composite strings joined by a \texttt{||} delimiter, with all components lowercased for normalization.

\paragraph{Web Search and Image Search.}
The primary key encodes both the search query and the question:
\begin{equation}
  \texttt{key} = \texttt{query} \;\|\;\texttt{question}
\end{equation}
If no question context is available, the key degrades to \texttt{query} alone. For inference trajectories a secondary key \texttt{query\textbar\textbar original} stores the raw observation independently of question context.

\paragraph{Web Read.}
The retrieved URL takes the role of the query:
\begin{equation}
  \texttt{key} = \texttt{url} \;\|\;\texttt{question}
\end{equation}
with fallback \texttt{url} when no question is present, and \texttt{url\textbar\textbar original} for the raw observation.

\paragraph{Reverse Image Search.}
This tool accepts both an image URL and an optional text refinement query, yielding a three-part key:
\begin{equation}
  \texttt{key} = \texttt{image\_url} \;\|\;\texttt{query} \;\|\;\texttt{question}
\end{equation}
Components are included only when non-empty, degrading gracefully to \texttt{image\_url\textbar\textbar query}, \texttt{image\_url\textbar\textbar question}, or \texttt{image\_url} alone.

\paragraph{OCR.}
The image source (URL or local path) serves as the primary identifier:
\begin{equation}
  \texttt{key} = \texttt{image\_source} \;\|\;\texttt{question}
\end{equation}
with fallback \texttt{image\_source} alone for raw observation storage.

This two-level scheme—question-contextualized summaries and raw observations—lets the training environment serve either a concise, relevance-filtered response or the full API output depending on what the rollout infrastructure requests.

% ------------------------------------------------------------------
\subsubsection{Response Extraction}
\label{appendix:cache:extraction}

\paragraph{From rollout files.}
Each \texttt{tool\_interact\_info} entry is identified by detecting the tool's XML tag in the \texttt{action} field. The query or URL is extracted from the tag's content (\emph{e.g.},\ \texttt{<text\_search\_text>query</text\_search\_text>}). For reverse image search, the tag content may encode both URL and query as \texttt{image\_url\textbar\textbar query}. The response is reconstructed by joining all non-empty strings in the \texttt{obs} list, stripping enclosing \texttt{<result>...</result>} tags, and splitting on the sentinel \texttt{"Response:"} to retain only the substantive output.

\paragraph{From trajectory files.}
Each trajectory step is inspected for the matching \texttt{action\_type} (\emph{e.g.},\ \texttt{"web\_search"}, \texttt{"ocr"}, \texttt{"reverse\_image\_search"}). The query or URL is read from \texttt{action\_parameters}, and the question is resolved from the top-level \texttt{question} field, falling back to nested \texttt{trajectory.question} or the first step's question. The \texttt{observation} and \texttt{observation\_summary} fields are then stored under their respective keys.

% ------------------------------------------------------------------
\subsubsection{Quality Filtering}
\label{appendix:cache:filtering}

Before inserting any entry, responses are validated to exclude API failures and empty results. A response is rejected if it:
\begin{itemize}[leftmargin=1.5em]
  \item is empty, \texttt{None}, or shorter than 10 characters;
  \item contains tool-specific error markers (\emph{e.g.},\ \textit{``search failed''}, \textit{``api error''}, \textit{``execution failed''}, \textit{``timeout''}, \textit{``rate limit exceeded''}, \textit{``quota exceeded''});
  \item contains semantically empty results (\emph{e.g.},\ \textit{``no search results found''}, \textit{``no image results''}, \textit{``no detailed information''}, \textit{``no content extracted''});
  \item begins with an error-class word (\textit{``error''}, \textit{``failed''}, \textit{``exception''}, \textit{``invalid''}, \textit{``empty''}) within its first three tokens.
\end{itemize}

Tool-specific exceptions apply where needed. For OCR, a response of \textit{``Text found in image: No text detected.''} is considered valid since it faithfully represents an image containing no text. For web search and reverse image search, an occurrence of \textit{``no results found''} embedded within a longer response ($>$50 characters) is not treated as a failure.

When multiple records share the same cache key, \textbf{the first valid response is retained} and subsequent duplicates are discarded, ensuring deterministic cache entries across training runs.

% ------------------------------------------------------------------
\subsection{Output Format}
\label{appendix:cache:format}

Each extractor emits a JSON file. For rollout-sourced caches the format is a flat dictionary:
\begin{lstlisting}[language=json, basicstyle=\ttfamily\small]
{
  "query||question": "Response text ...",
  "query":           "Response text ..."
}
\end{lstlisting}

For inference-sourced caches the output contains two sub-dictionaries:
\begin{lstlisting}[language=json, basicstyle=\ttfamily\small]
{
  "by_query_question": { "query||question": "observation_summary ..." },
  "by_query_original":  { "query||original": "raw observation ..." }
}
\end{lstlisting}
(with analogous structure for URL-based and image-URL-based tools).

At training time, the agent's tool server looks up incoming invocation parameters in the appropriate cache file. On a hit, the stored response is returned immediately without any external API call, eliminating the dominant source of latency and cost variability during policy-gradient rollouts.

\newpage
\section{Experiment Details}
\subsection{Human Validation Protocol}
\label{app:human_validation}

To assess the quality of the generated multi-hop questions, we conduct a human
annotation study on \datasetname~ covering five seed VQA datasets: FVQA, InfoSeek, InfoVQA,
News, and OK-VQA. Annotators evaluate each sample along two independent
dimensions.

\paragraph{Annotation Interface.}
We prepare a structured Excel workbook with one tab per benchmark dataset.
Each row corresponds to one multi-hop sample and contains the following
columns, color-coded by hop group for readability:

\begin{itemize}[noitemsep, topsep=4pt, leftmargin=1.5em]
  \item \textbf{id}: unique sample identifier.
  \item \textbf{question}: the final merged multi-hop question.
  \item \textbf{image url}: a clickable link to the source image (hosted on S3).
  \item \textbf{understand question?}: annotator response (True / False dropdown).
  \item For each hop $k = 1, \ldots, K$:
    \begin{itemize}[noitemsep, leftmargin=1.2em]
      \item \textbf{hop $k$ question}: the sub-question for this hop.
      \item \textbf{hop $k$ answer}: the expected answer for this hop.
      \item \textbf{hop $k$ url}: a clickable reference link (image URL for
            Hop~1; retrieved webpage URL for Hop~$k \geq 2$).
      \item \textbf{hop $k$ correct?}: annotator response (True / False dropdown).
    \end{itemize}
\end{itemize}

\noindent
All response columns use in-cell True/False dropdowns to standardize input and
minimise annotation error.

\paragraph{Task 1: Image Necessity Check.}
Prior to the comprehension check, annotators assess whether
the final multi-hop question \emph{genuinely requires} the image to answer. 
A question passes this check if the image provides the only means to determine the
specific answer---for instance, by identifying a person, landmark, logo, or
object that would be otherwise unspecified. 
Questions answerable from general knowledge alone, without reference to the image, are flagged.

\paragraph{Task 2: Overall Question Comprehension.}
Annotators first read the final multi-hop question and open the image URL.
They assess whether the question is grammatically sound, logically coherent,
and unambiguous given the image. A question is marked \textsc{True} if its
intent is clearly understandable---even if the phrasing is not perfectly
natural---and \textsc{False} if it is confusing, contradictory, or
uninterpretable without additional context.

\paragraph{Task 3: Per-Hop Factual Accuracy.}
For each hop, annotators verify the correctness of the (sub-question, answer)
pair against the provided reference. Specifically:

\begin{itemize}[noitemsep, topsep=4pt, leftmargin=1.5em]
  \item \textbf{Hop 1} is answered from the image; annotators open the image
        URL and check whether the stated answer accurately describes the visual
        content.
  \item \textbf{Hops $k \geq 2$} are answered from retrieved webpages;
        annotators open the hop URL and verify whether the answer is factually
        supported by the page content.
\end{itemize}

A hop is marked \textsc{True} if the sub-question and answer are factually
correct, supported by the reference, and logically consistent with the
preceding hop. It is marked \textsc{False} if the answer is incorrect,
hallucinated, or breaks the reasoning chain.

\subsection{Search Agent Implementation Details}
\label{app:search_agent}

\subsubsection{Agent Architecture Overview}

Our multimodal search agent follows the \textbf{ReAct} (Reasoning and Acting) paradigm, interleaving natural-language reasoning steps with grounded tool executions to iteratively gather information before producing a final answer. The agent is implemented as an abstract base class (\texttt{MultimodalDeepResearchTesterBase}) that supports different interchangeable inference backends such as  OpenAI GPT-5, vLLM (for locally-served open-weight models), and Google Gemini. All backends share the same tool interface, prompt templates, and loop logic.

\subsubsection{Tool Registry}
\label{app:tool_agent}

The agent exposes a fixed registry of seven tools. Each tool is defined with a natural-language description, typed parameters, an XML-style invocation format, and a usage example. The active tool set is configurable per run; unused tools are omitted from the system prompt.

\begin{table}[h]
\centering
\small
\setlength\tabcolsep{0.3pt}
\begin{tabular}{lll}
\toprule
\textbf{Tool Name} & \textbf{XML Tag} & \textbf{Purpose} \\
\midrule
\texttt{web\_text\_to\_text\_search}  & \texttt{<text\_search\_text>}  & Text-to-text web search; returns snippets and URLs \\
\texttt{web\_text\_to\_img\_search}   & \texttt{<text\_search\_image>} & Text-to-image web search; returns image descriptions \\
\texttt{web\_url\_reader}             & \texttt{<web\_read>}           & Full-page content extraction from a URL \\
\texttt{web\_image\_to\_text}         & \texttt{<image\_search\_text>} & Google Lens reverse-image search; identifies objects \\
\texttt{ocr\_tool}                    & \texttt{<ocr\_tool>}           & OCR extraction of all visible text from an image \\
\texttt{ipython\_code}                & \texttt{<python>}              & IPython code execution with persistent state \\
% \texttt{bash\_terminal}               & \texttt{<bash>}                & Bash command execution \\
\bottomrule
\end{tabular}
\caption{Tool registry available to the multimodal search agent.}
\label{tab:tool_registry}
\end{table}

\subsubsection{ReAct Loop}
\label{app:react}

The agent runs up to \textbf{6 iterations} of the Reason--Act--Observe cycle. The cycle is implemented via a stateful chat message list that grows with each step.

\paragraph{Step 1 -- Initialization.}
The chat history is seeded with a system prompt that includes the tool registry descriptions and stopping criteria. On the first iteration, the research question and the associated image (base64-encoded inline) are included in the first user message.

\paragraph{Step 2 -- Reasoning.}
The model receives the full chat history and produces a structured JSON response:

\begin{verbatim}
{
  "reasoning": "...",
  "action": {
    "action_type": "web_search",
    "action_description": "...",
    "action_parameters": {"query": "..."},
    "expected_outcome": "..."
  },
  "should_stop": false,
  "confidence": 0.8
}
\end{verbatim}

The JSON is parsed by scanning for the first balanced \texttt{\{...\}} substring, falling back to a default web-search action on parse failure.

\paragraph{Step 3 -- Reflection.}
Before executing the action, a reflection sub-step can be enabled. The model is prompted to critique the proposed action---evaluating query quality, redundancy with prior steps, and tool choice---and may modify the \texttt{action\_parameters} or switch to a different tool. 

\paragraph{Step 4 -- Action Execution.}
The action JSON is translated into an XML-tagged string (e.g., \texttt{<text\_search\_text>query</text\_search\_text>}) and dispatched as an HTTP \texttt{POST} request to a tool server at \texttt{/get\_observation}. The server returns a JSON observation. URL placeholders (\texttt{[IMAGE]}) in action parameters are resolved to the actual image URL at dispatch time.

\paragraph{Step 5 -- Observation.}
The raw server response is either used directly or compressed by a GPT-5 summarizer (one prompt per tool type) and appended to the chat history as a user message. 

\paragraph{Step 6 -- Stopping.}
The loop exits when: (a) the model sets \texttt{should\_stop: true} with \texttt{confidence} $> 0.7$; (b) two consecutive tool calls fail; (c) the estimated context length exceeds the configured token limit (the last ReAct cycle is removed and the agent proceeds directly to the final answer); or (d) the maximum of 6 iterations is reached.

\subsubsection{Context Management}

Token length is estimated as $\lfloor \text{total characters} / 4 \rfloor$ for GPT and Gemini backends, and by the vLLM tokenizer for local models. Image content is approximated at 256 tokens per image. When the estimate exceeds \texttt{max\_context\_tokens} (default 128k), the most recent full ReAct cycle---observation turn, reasoning turn, and reflection messages if enabled---is removed from the chat history before the final answer step.

\subsubsection{Final Answer Generation}

After the loop terminates, the model is prompted to synthesize its findings and produce a concise answer enclosed in a \LaTeX\ box:

\begin{quote}
\textit{Based on the research findings provided, provide a direct answer to the research question. $\ldots$ Put your final concise answer inside \texttt{\textbackslash boxed\{\}}.}
\end{quote}

The boxed expression is subsequently extracted for automatic evaluation.

\subsection{LLM-as-Judge Evaluation}
\label{app:llm-judge}

We evaluate agent trajectories using \texttt{check\_answer\_correctness}, which calls GPT-5 as a judge to perform a binary correctness check after each trajectory completes. The judge receives the following prompt:

\begin{tcolorbox}[colback=gray!8, colframe=gray!50,
  title=\textbf{Prompt: Answer Correctness Check (GPT-5)},
  fonttitle=\small\bfseries, ]
\small
\textbf{System:} You are an expert answer evaluator. Respond only with valid JSON.

\medskip
\textbf{User:} You are an expert answer evaluator. Compare the model's answer to the ground truth and determine if they are semantically equivalent.

\medskip
\textbf{Question:} \{question\}

\textbf{Ground Truth Answer:} \{ground\_truth\}

\textbf{Model's Answer:} \{final\_answer\}

\medskip
\textbf{Instructions:}
\begin{itemize}
    \item The model's answer is correct if it conveys the same core information as the ground truth, even if worded differently.
    \item Minor differences in phrasing, formatting, or additional context are acceptable if the key answer is correct.
    \item Look for the actual answer, which might be inside \texttt{\textbackslash boxed\{\}}.
    \item Additionally, check if the ground truth answer appears anywhere in the model's reasoning/explanation outside of \texttt{\textbackslash boxed\{\}}. This indicates correct reasoning even if the final answer format is different.
\end{itemize}

\medskip
Respond with a JSON object:\\
\texttt{\{"is\_correct": true/false, "is\_correct\_reasoning": true/false, "reason": "brief explanation"\}}

\medskip
\textbf{Where:}
\begin{itemize}
    \item \texttt{"is\_correct"}: true if the answer (especially in \texttt{\textbackslash boxed\{\}}) matches the ground truth.
    \item \texttt{"is\_correct\_reasoning"}: true if the ground truth can be found in the reasoning/explanation outside of \texttt{\textbackslash boxed\{\}}, even if \texttt{"is\_correct"} is false.
\end{itemize}
\end{tcolorbox}

The judge returns two boolean fields:

\begin{itemize}
    \item \textbf{\texttt{is\_correct}}: \texttt{true} if the content enclosed in \texttt{\textbackslash boxed\{\}} is semantically equivalent to the ground truth, allowing for minor differences in phrasing or formatting.
    \item \textbf{\texttt{is\_correct\_reasoning}}: \texttt{true} if the ground-truth answer appears anywhere in the model's reasoning or explanation \emph{outside} of \texttt{\textbackslash boxed\{\}}, even when \texttt{is\_correct} is \texttt{false}. This distinguishes cases where the model arrived at the correct answer through valid reasoning but failed to format it correctly in the final box.
\end{itemize}

\section{More Study}
\subsection{Ablation of Data generation pipelines}
\label{app:ablation_data}

We iterate the desgin of data generation pipelines multiple times, each time, our 4 authors label $100$ generation QA and update the prompts and design.

We first identify an issue called \textbf{image redundancy}. During merging, specific entities may make a question answerable without the image. We find that $14\%$ of samples suffer from this issue.  
For example, consider the merged question: ``What organization was established to coordinate the policies and operations of the United States representatives on the Fund and the Bank in the country where this shop's official currency was first recognized as a currency of the world?'' Here, the image only depicts the United States, so the question can be answered without it.  
To address this, we add a validation step: GPT-5 is asked to answer the question \textit{without} the image. If the answer matches the one obtained with the image, the image is considered redundant. After applying this step, the redundancy rate drops to $6\%$.

We also observe that $15\%$ of samples suffer from \textbf{inconsistent entities}. Even when the entity name is the same, the actual entity may differ. 
For example, ``Newark'' in one answer refers to Newark, CA, while in the next question it refers to Newark, NJ. 
To address this, we add descriptive context to each entity, which is then used in subsequent question generation. For instance, ``Newark, a city in CA, US'' ensures consistent reference across questions. 
After applying this, the inconsistency rate drops to $3\%$.
In addition, we find that $12\%$ of questions suffer from \textbf{contextual dependency}: these questions are not standalone, often referring to ``provided content,'' ``context,'' or ``mentions'' within the task rather than forming natural, independent questions. We address this by adding prompts during question generation and validation. After this step, the rate of contextual dependency drops to $2\%$.

\subsection{Dataset-Specific Tool Use and Search Depth Analysis}
\label{app:tool_turn_analysis}

We present per-dataset breakdowns of tool usage and turn distributions for all five benchmarks evaluated in this work. For each benchmark, we compare three settings: the base model before RL training (Before Training), the model after RL training (After Training), and GPT-5 as an oracle reference. Tool usage is measured as the percentage of scenarios in which each tool is invoked at least once. Turn distribution shows the percentage of trajectories terminating at each number of search steps.

\begin{figure}[tb!]
\centering
\begin{minipage}{0.45\textwidth}
    \centering
    \includegraphics[width=\linewidth]{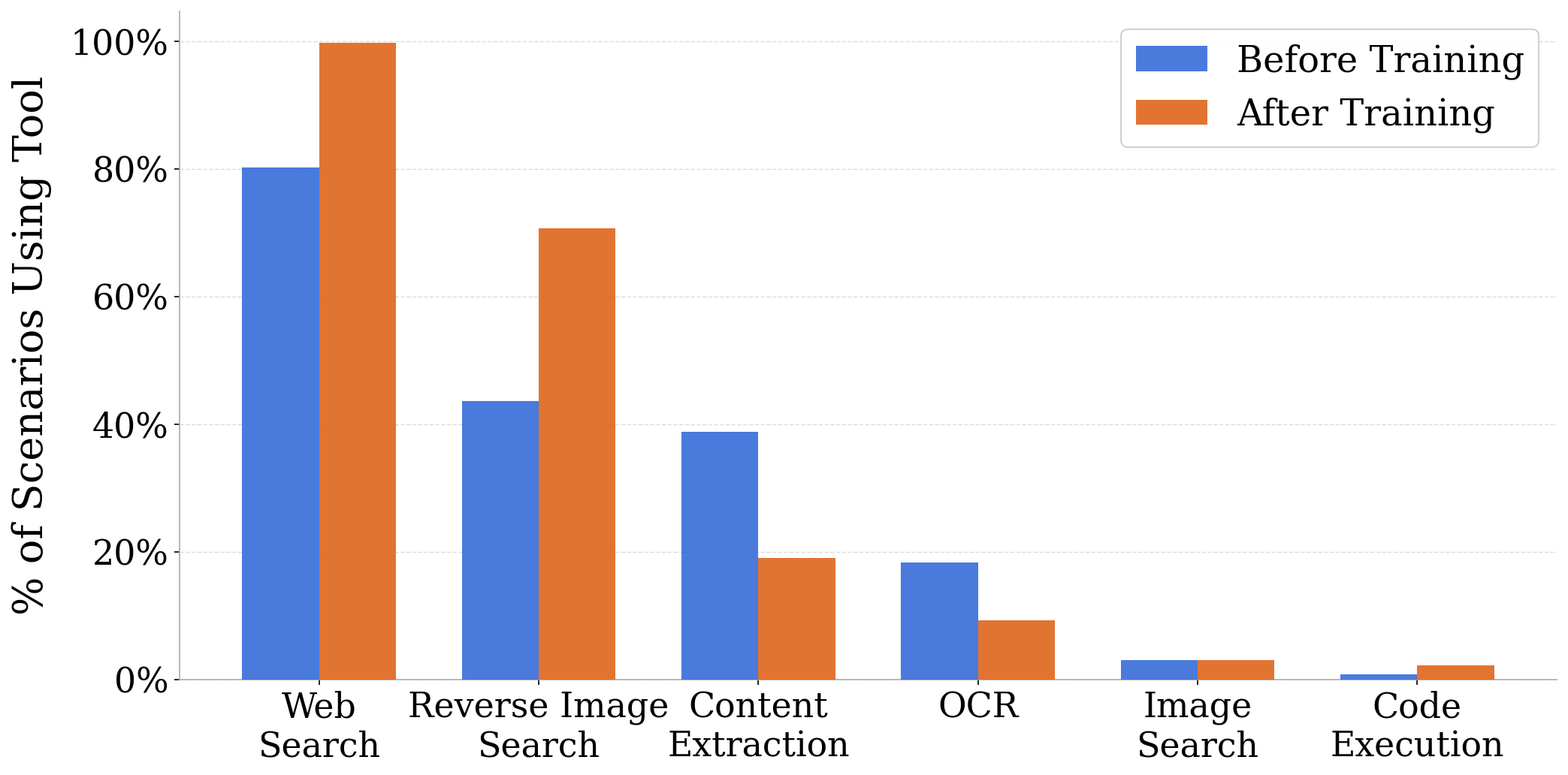}
    \caption{Tool usage distribution on BrowseComp-VL.}
    \label{fig:browsecomp_tool}
\end{minipage}
\hfill
\begin{minipage}{0.45\textwidth}
    \centering
    \includegraphics[width=\linewidth]{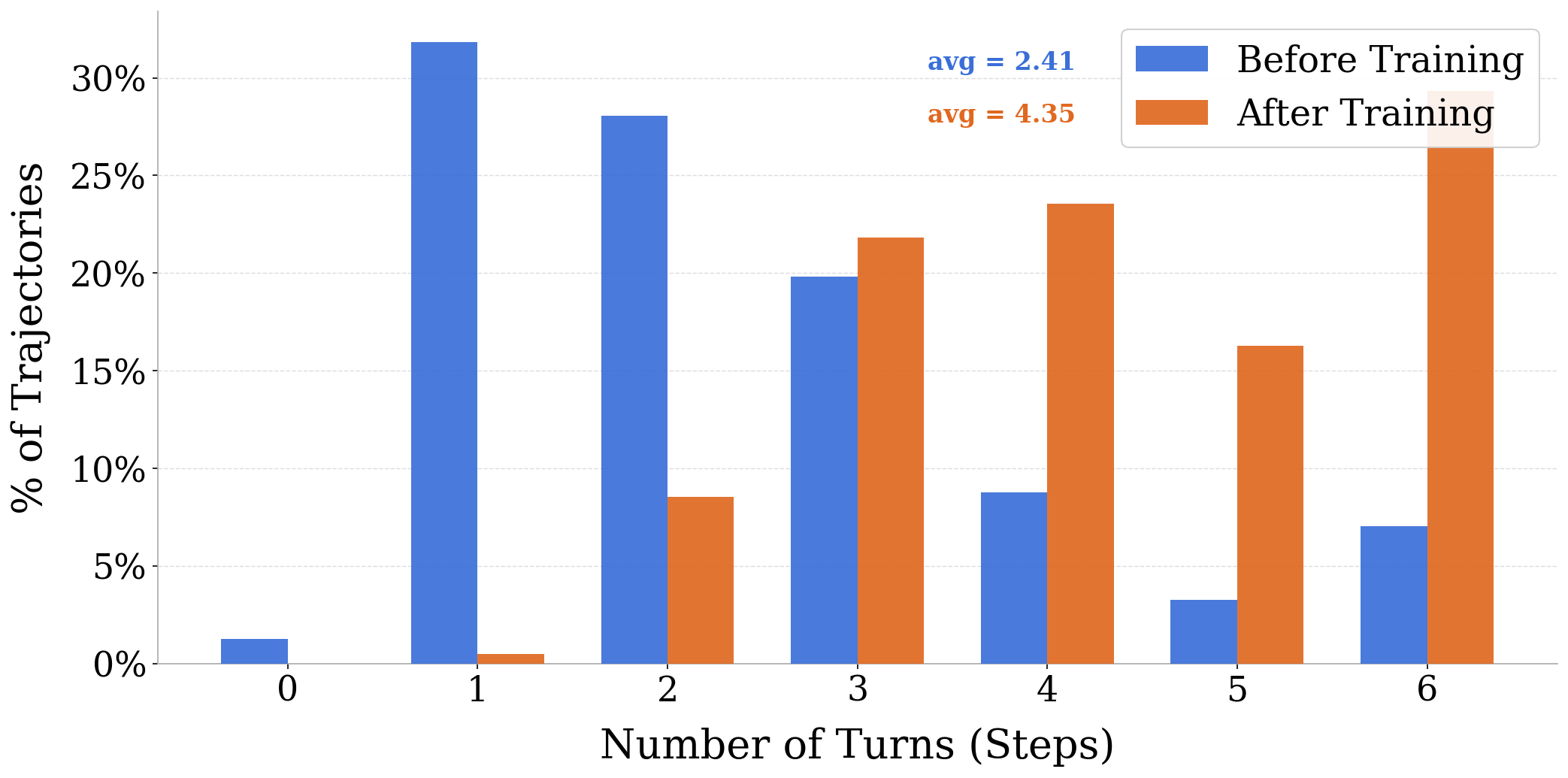}
    \caption{Turn distribution on BrowseComp-VL.}
    \label{fig:browsecomp_turn}
\end{minipage}
\end{figure}

\begin{figure}[tb!]
\centering
\begin{minipage}{0.45\textwidth}
    \centering
    \includegraphics[width=\linewidth]{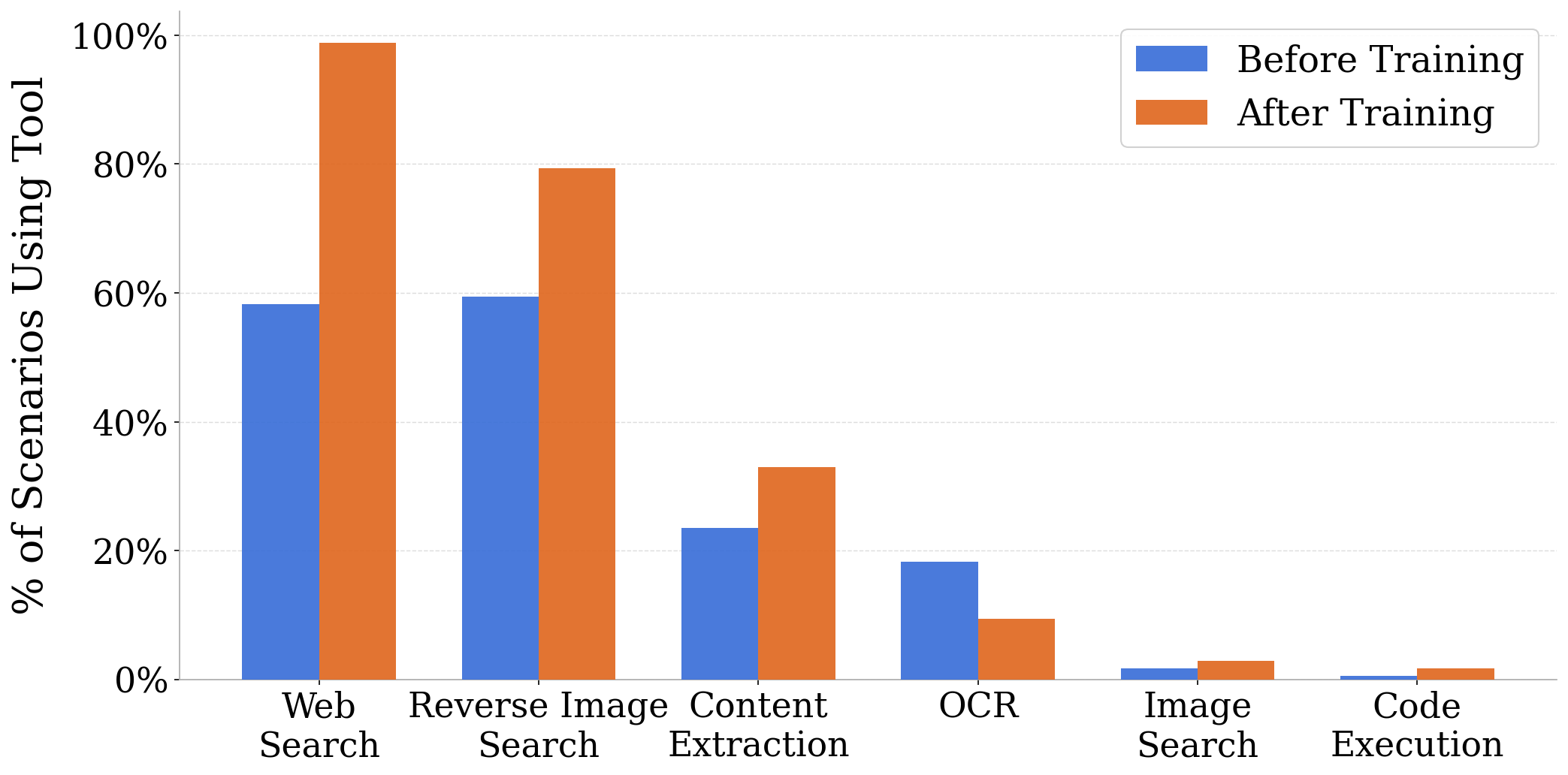}
    \caption{Tool usage distribution on MMSearch-VL.}
    \label{fig:mmsearch_tool}
\end{minipage}
\hfill
\begin{minipage}{0.45\textwidth}
    \centering
    \includegraphics[width=\linewidth]{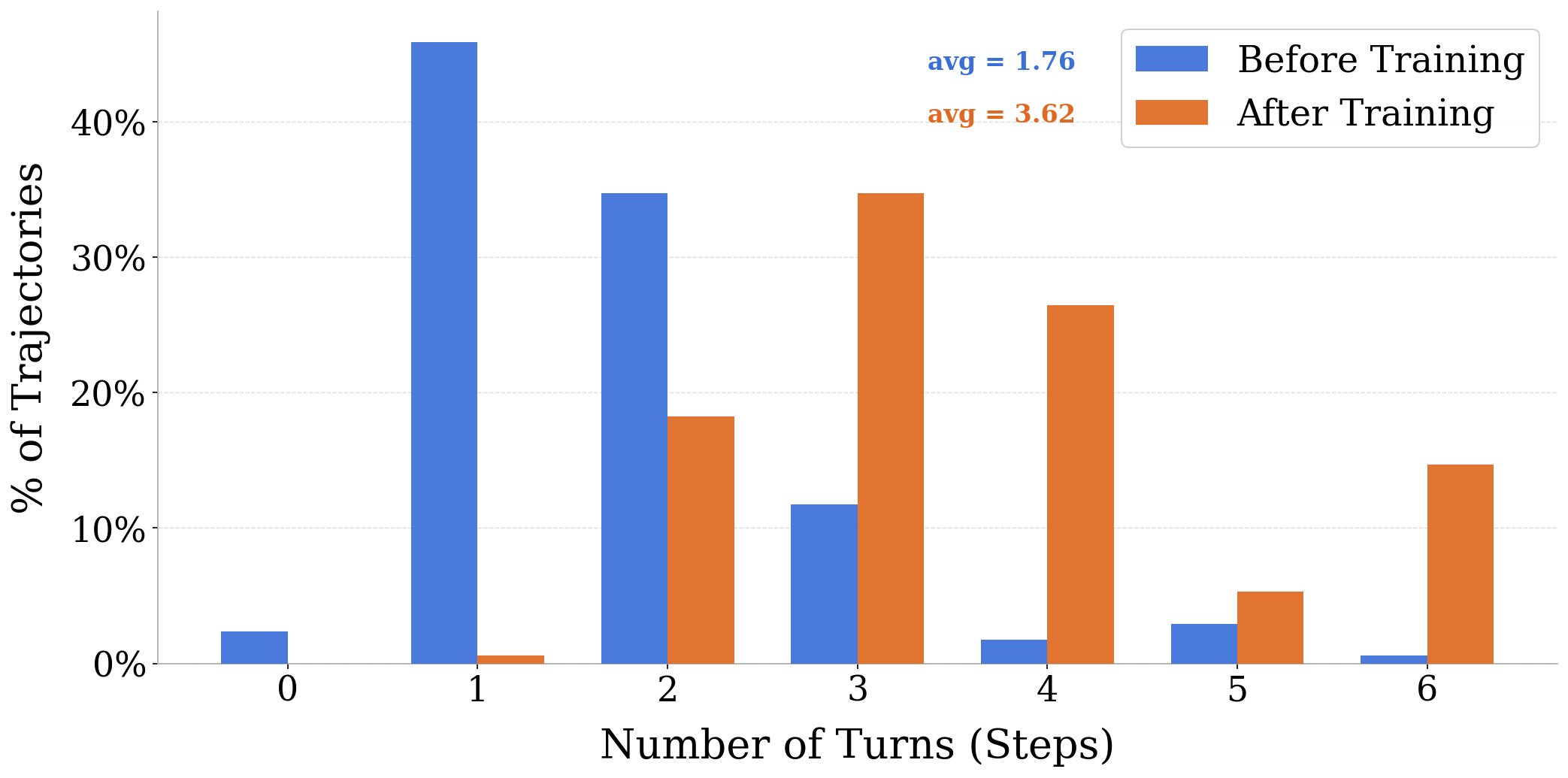}
    \caption{Turn distribution on MMSearch-VL.}
    \label{fig:mmsearch_turn}
\end{minipage}
\end{figure}

\begin{figure}[tb!]
\centering
\begin{minipage}{0.45\textwidth}
    \centering
    \includegraphics[width=\linewidth]{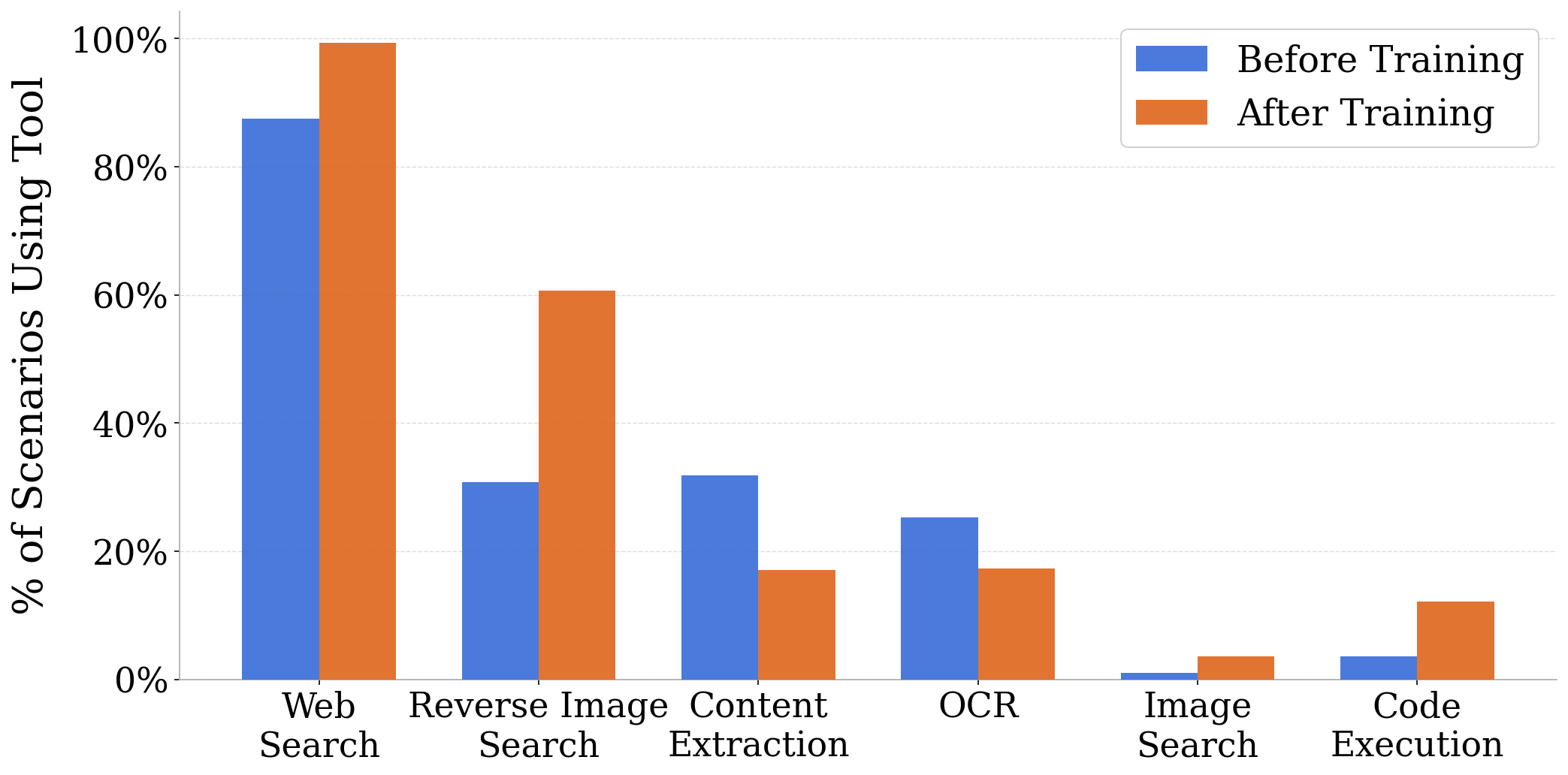}
    \caption{Tool usage distribution on HR-MMSearch-VL.}
    \label{fig:hr_mmsearch_tool}
\end{minipage}
\hfill
\begin{minipage}{0.45\textwidth}
    \centering
    \includegraphics[width=\linewidth]{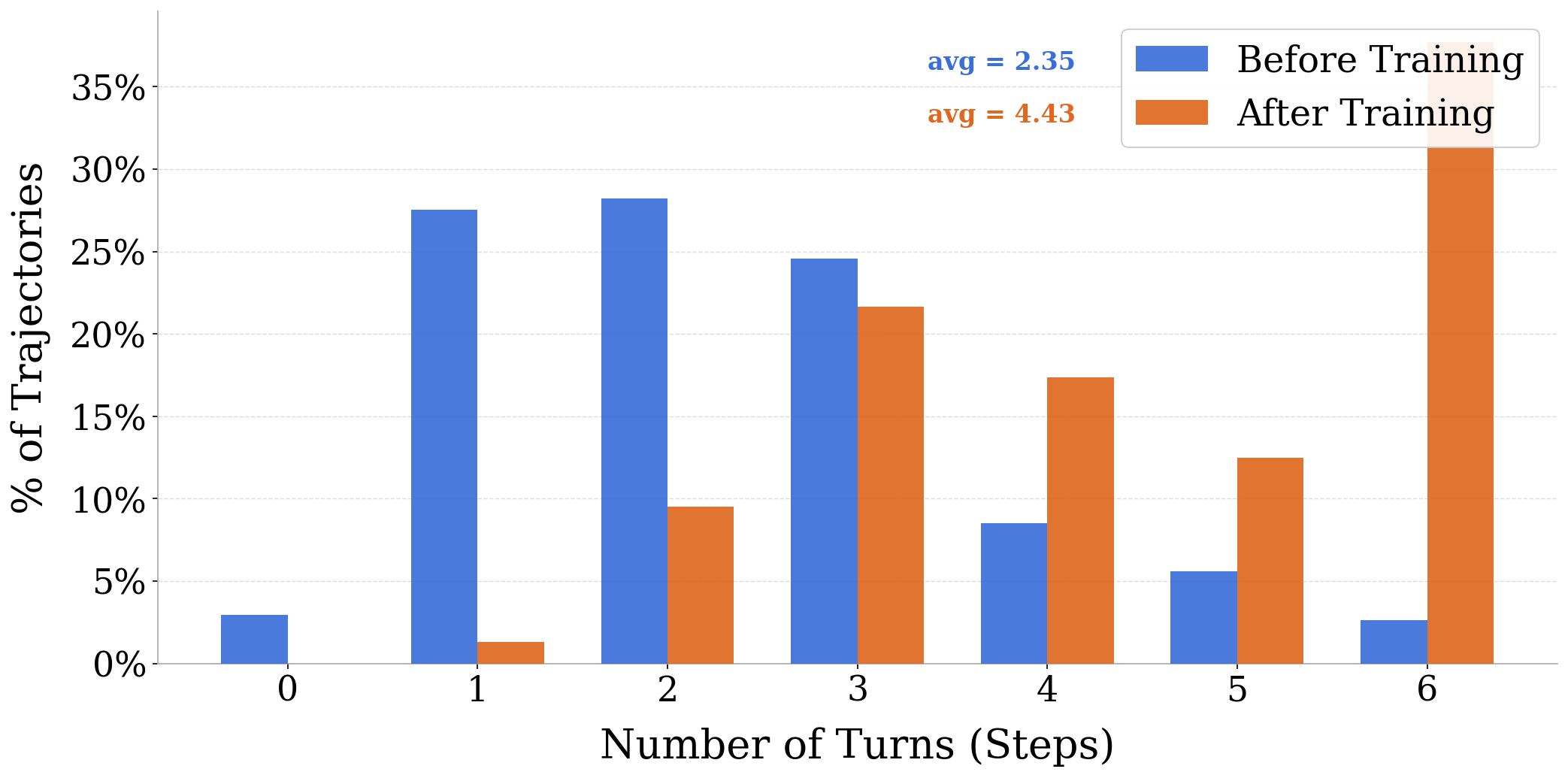}
    \caption{Turn distribution on HR-MMSearch-VL.}
    \label{fig:hr_mmsearch_turn}
\end{minipage}
\end{figure}

\begin{figure}[tb!]
\centering
\begin{minipage}{0.45\textwidth}
    \centering
    \includegraphics[width=\linewidth]{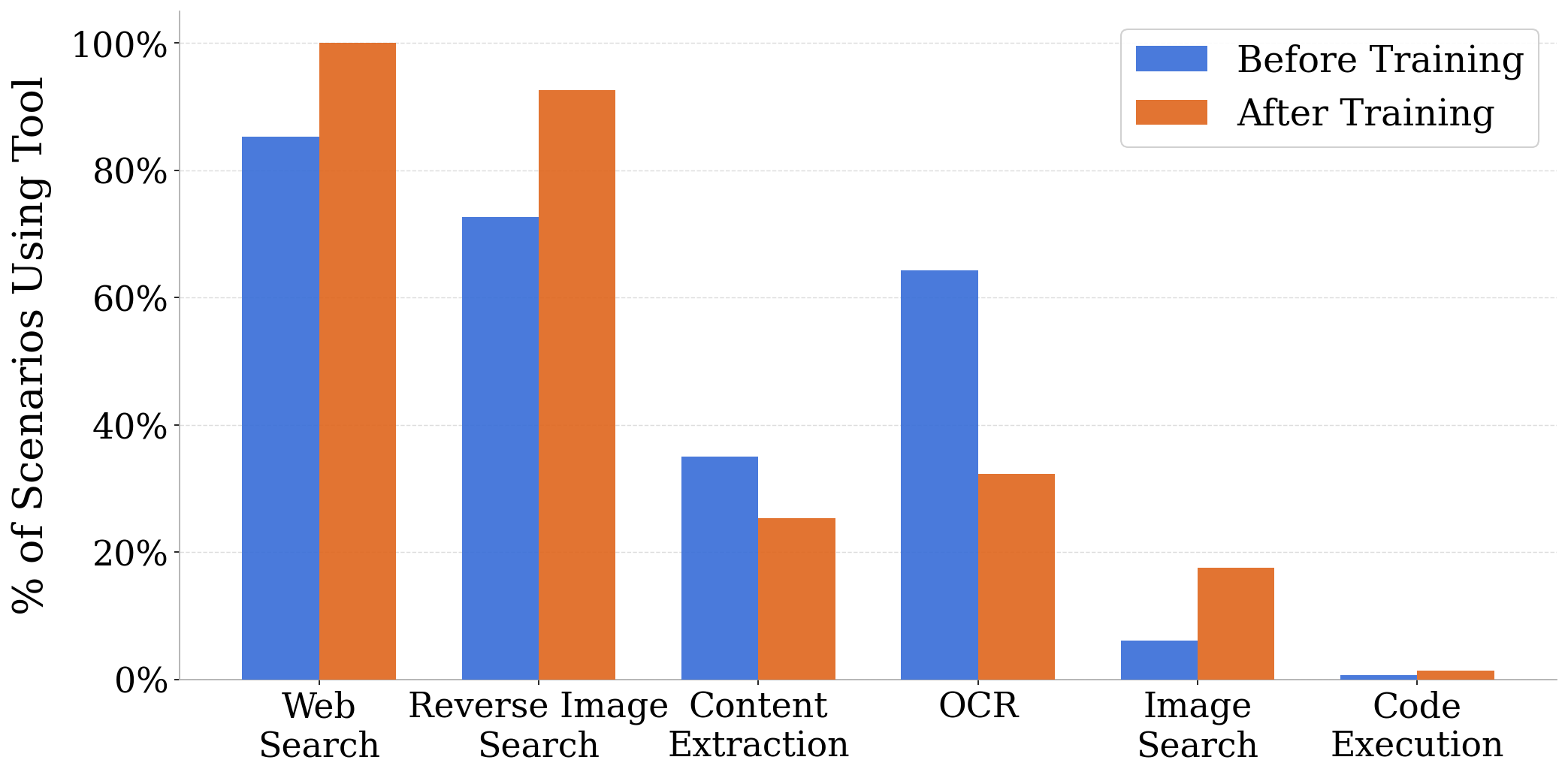}
    \caption{Tool usage distribution on MMSearch-Plus.}
    \label{fig:mmsearch_plus_tool}
\end{minipage}
\hfill
\begin{minipage}{0.45\textwidth}
    \centering
    \includegraphics[width=\linewidth]{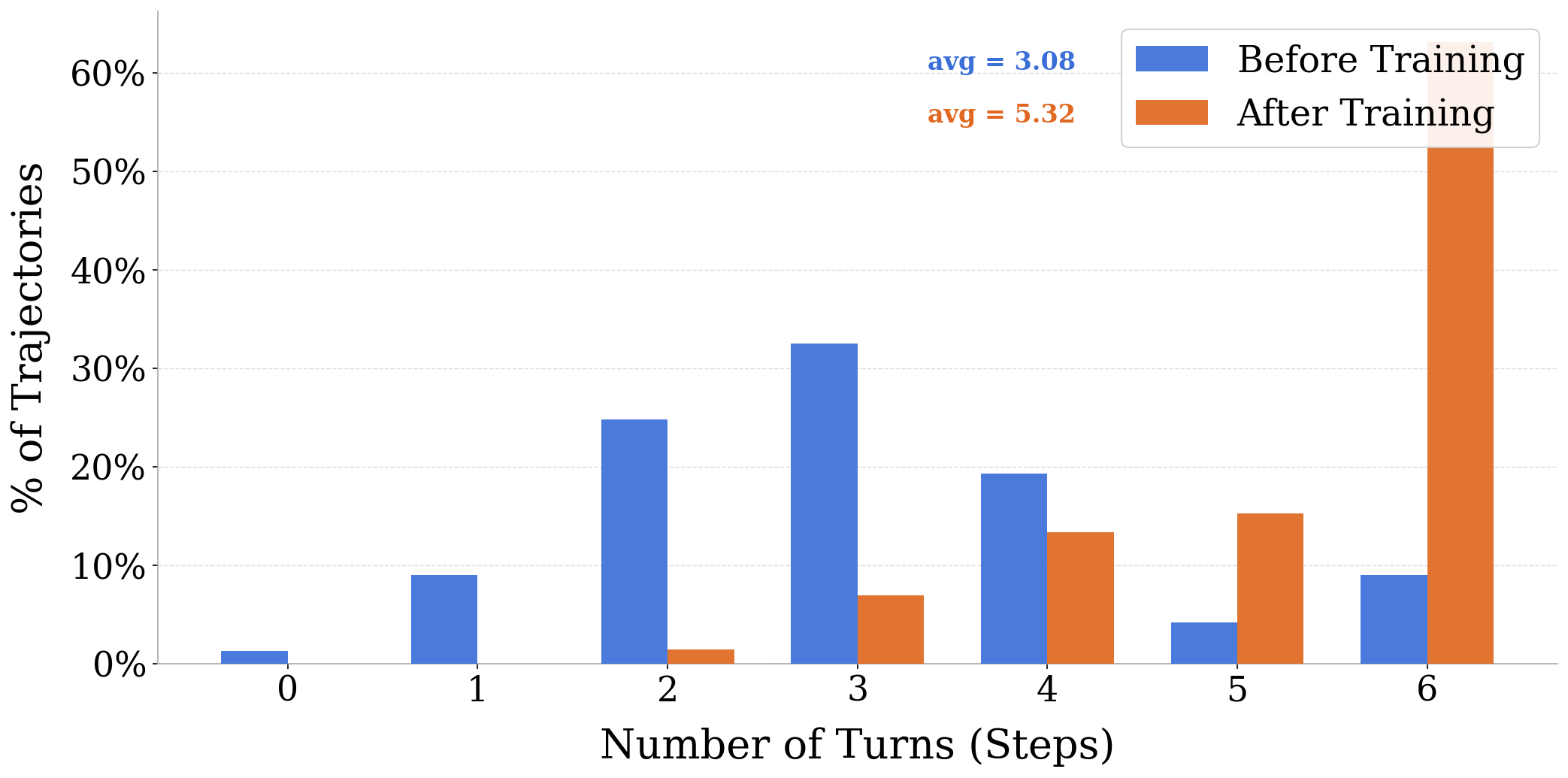}
    \caption{Turn distribution on MMSearch-Plus.}
    \label{fig:mmsearch_plus_turn}
\end{minipage}
\end{figure}

\begin{figure}[tb!]
\centering
\begin{minipage}{0.45\textwidth}
    \centering
    \includegraphics[width=\linewidth]{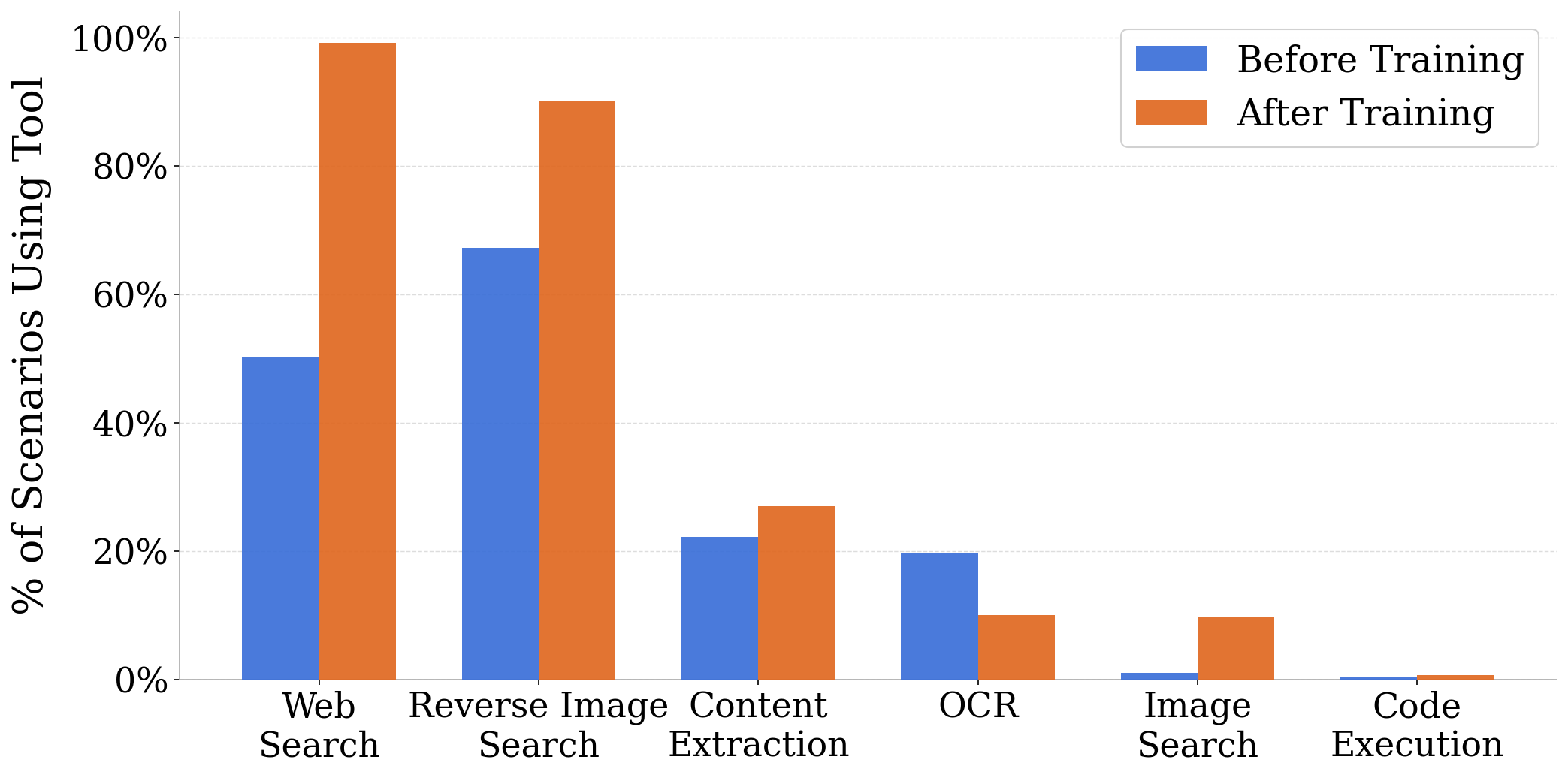}
    \caption{Tool usage distribution on FVQA.}
    \label{fig:fvqa_tool}
\end{minipage}
\hfill
\begin{minipage}{0.45\textwidth}
    \centering
    \includegraphics[width=\linewidth]{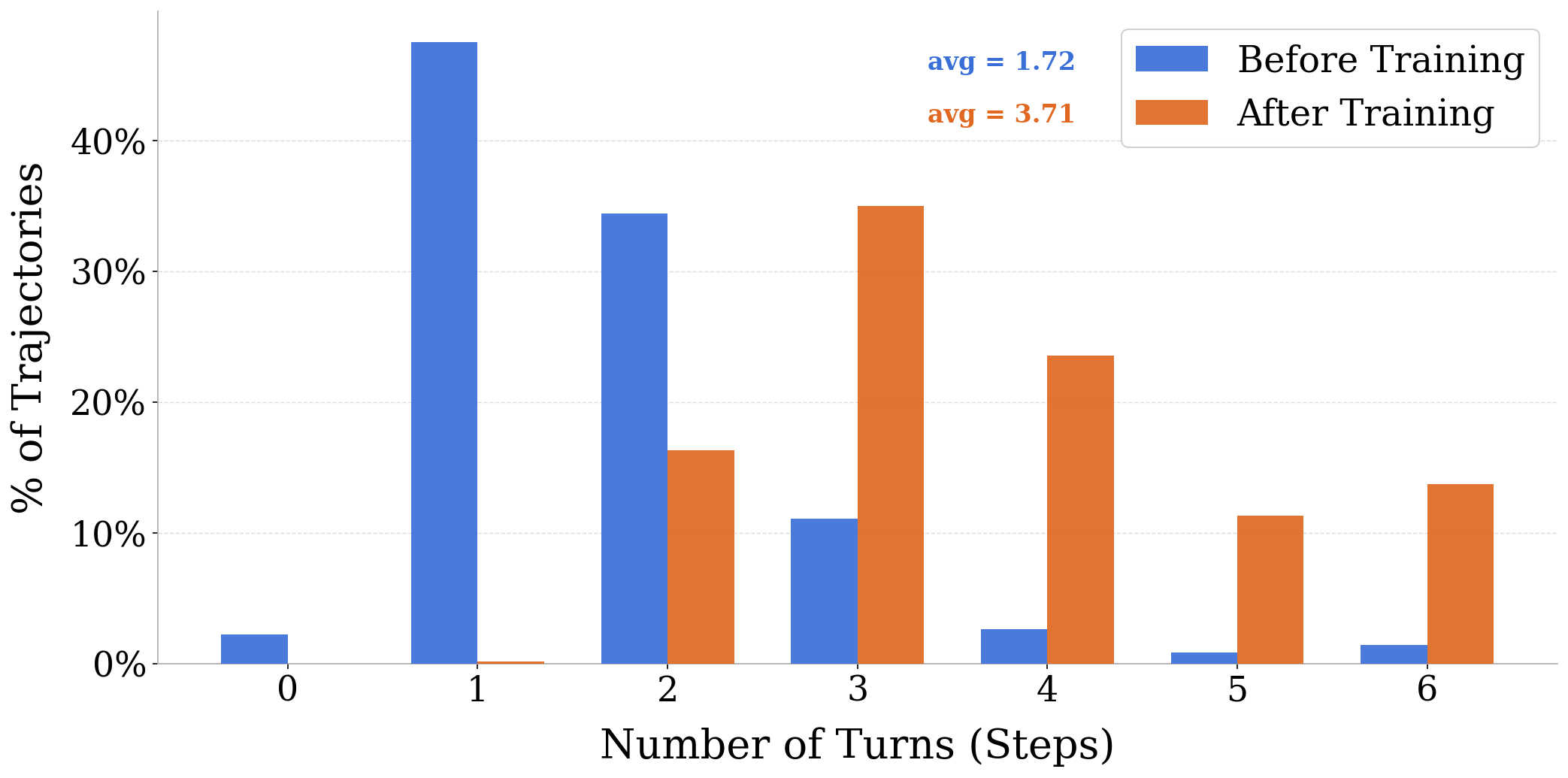}
    \caption{Turn distribution on FVQA.}
    \label{fig:fvqa_turn}
\end{minipage}
\end{figure}

\clearpage
\section{Examples}
\label{app:examples}

\subsection*{Example 1}
\begin{figure}[b!]
    \centering
    \includegraphics[width=0.5\linewidth]{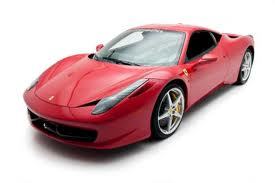}
    \caption{Image for Example 1.}
    \label{fig:Ferrari}
\end{figure}

\noindent\textbf{Image:} See Figure~\ref{fig:Ferrari}.
\vspace{0.5em}

\noindent\textbf{Individual Hops:}

\begin{enumerate}
    \item \textbf{Q (Image):} What is the brand of the vehicle in the image? \\
          \textbf{A:} Ferrari
    \item \textbf{Q:} Who founded Ferrari? \\
          \textbf{A:} Enzo Ferrari
    \item \textbf{Q:} Which flying ace's prancing horse emblem did Enzo Ferrari adopt for his cars? \\
          \textbf{A:} Francesco Baracca
    \item \textbf{Q:} Who was the sculptor of the monument dedicated to Francesco Baracca? \\
          \textbf{A:} Domenico Rambelli
    \item \textbf{Q:} In which city is Domenico Rambelli's monument to Francesco Baracca located? \\
          \textbf{A:} Lugo
\end{enumerate}

\noindent\textbf{Merged Questions (step-by-step composition):}

\begin{enumerate}
    \item\textit{After Hop 1$\to$2:} Who founded the brand of the vehicle in the image?
    \item\textit{After Hop 2$\to$3:} Which flying ace's prancing horse emblem was adopted for the cars of the brand of the vehicle in the image?
    \item\textit{After Hop 3$\to$4:} Which sculptor created the monument dedicated to the flying ace whose prancing horse emblem was later adopted for the cars of the brand of the vehicle in the image?
    \item\textit{After Hop 4$\to$5 (Final):} In which city is the monument located that honors the flying ace whose prancing horse emblem was later adopted for the cars of the brand of the vehicle in the image?
\end{enumerate}
\noindent\textbf{Final Answer:} \textit{Lugo}

\clearpage
\subsection*{Example 2}
\begin{figure}[b!]
    \centering
    \includegraphics[width=0.3\linewidth]{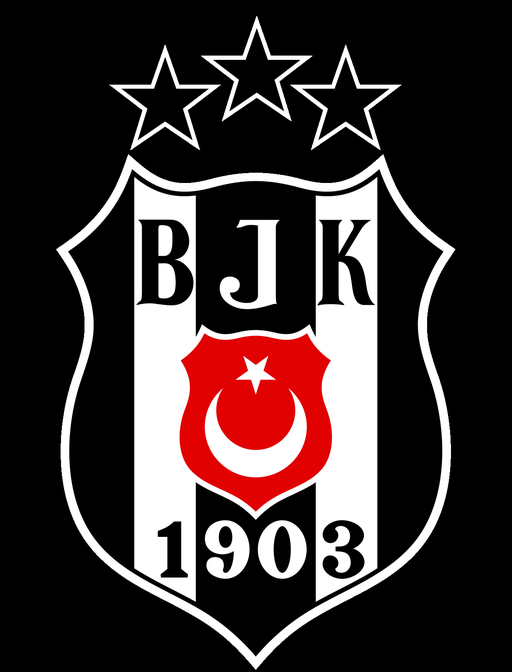}
    \caption{Image for Example 2.}
    \label{fig:BJK}
\end{figure}

\noindent\textbf{Image:} See Figure~\ref{fig:BJK}.

\noindent\textbf{Individual Hops:}

\begin{enumerate}
    \item \textbf{Q (Image):} What organization is represented by the emblem in the image? \\
          \textbf{A:} Be\c{s}ikta\c{s} J.K.
    \item \textbf{Q:} In which city is Be\c{s}ikta\c{s} J.K.'s home stadium located? \\
          \textbf{A:} Istanbul
    \item \textbf{Q:} What is the name of the underground metro line in Istanbul that is considered the world's second-oldest? \\
          \textbf{A:} Istanbul T\"{u}nel
    \item \textbf{Q:} Which sultan granted the concession to build the Istanbul T\"{u}nel? \\
          \textbf{A:} Sultan Abd\"{u}laziz
    \item \textbf{Q:} Which British monarch invested Sultan Abd\"{u}laziz with the Order of the Garter? \\
          \textbf{A:} Queen Victoria
\end{enumerate}

\noindent\textbf{Merged Questions (step-by-step composition):}

\begin{enumerate}
    \item \textit{After Hop 1$\to$2:} In which city is the home stadium of the organization represented by the emblem in the image located?
    \item \textit{After Hop 2$\to$3:} What is the name of the underground metro line in the city where the home stadium of the organization represented by the emblem in the image is located, which is considered the world's second-oldest?
    \item \textit{After Hop 3$\to$4:} Which sultan granted the concession to build the underground metro line in the city where the home stadium of the organization represented by the emblem in the image is located, which is considered the world's second-oldest?
    \item \textit{After Hop 4$\to$5 (Final):} Which British monarch invested with the Order of the Garter the sultan who granted the concession to build the underground metro line in the city where the home stadium of the organization represented by the emblem in the image is located, which is considered the world's second-oldest?
\end{enumerate}

\noindent\textbf{Final Answer:} \textit{Queen Victoria}

\clearpage
\subsection*{Example 3}

\noindent\textbf{Image:} See Figure~\ref{fig:Stand-on-Zanzibar}.

\begin{figure}
    \centering
    \includegraphics[width=0.4\linewidth]{figures/examples/153.jpg}
    \caption{Image for Example 3.}
    \label{fig:Stand-on-Zanzibar}
\end{figure}

\noindent\textbf{Individual Hops:}

\begin{enumerate}
    \item \textbf{Q (Image):} Which book in the image predicted the fall of Detroit into poverty? \\
          \textbf{A:} \textit{Stand on Zanzibar}

    \item \textbf{Q:} Who wrote \textit{Stand on Zanzibar}? \\
          \textbf{A:} John Brunner

    \item \textbf{Q:} In which Scottish city did John Brunner die? \\
          \textbf{A:} Glasgow

    \item \textbf{Q:} Which saint is associated with giving Glasgow its name meaning `dear green place'? \\
          \textbf{A:} Saint Mungo

    \item \textbf{Q:} In which Scottish town was Saint Mungo born? \\
          \textbf{A:} Culross
\end{enumerate}

\noindent\textbf{Merged Questions (step-by-step composition):}

\begin{enumerate}
    \item \textit{After Hop 1$\to$2:} Who wrote the book in the image that predicted the fall of Detroit into poverty?

    \item \textit{After Hop 2$\to$3:} In which Scottish city did the author of the book in the image that predicted the fall of Detroit into poverty die?

    \item \textit{After Hop 3$\to$4:} Which saint is associated with giving the name meaning `dear green place' to the Scottish city where the author of the book in the image that predicted the fall of Detroit into poverty died?

    \item \textit{After Hop 4$\to$5 (Final):} In which Scottish town was the saint born who is associated with giving the name meaning `dear green place' to the Scottish city where the author of the book in the image that predicted the fall of Detroit into poverty died?
\end{enumerate}

\noindent\textbf{Final Answer:} \textit{Culross}

\clearpage
\subsection*{Example 4}

\begin{figure}[b!]
    \centering
    \includegraphics[width=0.3\linewidth]{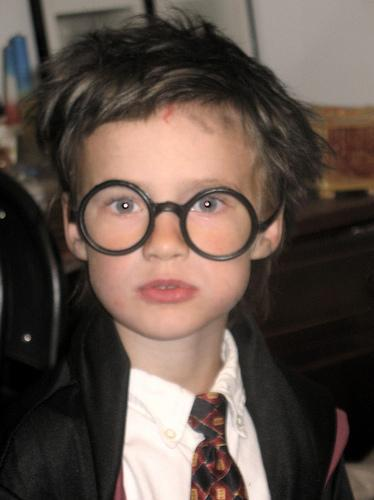}
    \caption{Image for Example 4.}
    \label{fig:Halloween}
\end{figure}

\noindent\textbf{Image:} See Figure~\ref{fig:Halloween}.

\noindent\textbf{Individual Hops:}

\begin{enumerate}
    \item \textbf{Q (Image):} What holiday is the boy in the image likely celebrating? \\
          \textbf{A:} Halloween

    \item \textbf{Q:} Which city is known as the Halloween Capital of the World? \\
          \textbf{A:} Anoka, Minnesota

    \item \textbf{Q:} From which lake does the Rum River in Anoka, Minnesota flow out? \\
          \textbf{A:} Mille Lacs Lake

    \item \textbf{Q:} Which U.S.\ president issued the executive order that first placed Spirit Island in Mille Lacs Lake under federal protection? \\
          \textbf{A:} Woodrow Wilson

    \item \textbf{Q:} On which transport ship did Woodrow Wilson travel to the peace negotiations after World War I? \\
          \textbf{A:} USS George Washington
\end{enumerate}

\noindent\textbf{Merged Questions (step-by-step composition):}

\begin{enumerate}
    \item \textit{After Hop 1$\to$2:} Which city is known as the capital of the holiday the boy in the image is likely celebrating?

    \item \textit{After Hop 2$\to$3:} From which lake does the river flowing out of the city known as the capital of the holiday the boy in the image is likely celebrating originate?

    \item \textit{After Hop 3$\to$4:} Which U.S.\ president issued the executive order that first placed Spirit Island in the lake from which the river flowing out of the city known as the capital of the holiday the boy in the image is likely celebrating originates, under federal protection?

    \item \textit{After Hop 4$\to$5 (Final):} On which transport ship did the U.S.\ president who first placed Spirit Island in the lake from which the river flowing out of the city known as the capital of the holiday the boy in the image is likely celebrating originates, travel to the peace negotiations after World War I?
\end{enumerate}

\noindent\textbf{Final Answer:} \textit{USS George Washington}

\end{document}